%% file: draft.tex
\documentclass[sigconf,authorversion=true]{acmart}

\usepackage{booktabs} % For formal tables

\usepackage{graphicx}
\usepackage{subfig}
\usepackage[linesnumbered, vlined, ruled]{algorithm2e}

\usepackage{xr}
\def\vector#1{\mbox{\boldmath $#1$}}
\newcommand{\CR}{C}

\setcopyright{rightsretained}
\copyrightyear{2020}
\acmYear{2020}
\setcopyright{acmcopyright}
\acmConference[GECCO '20]{Genetic and Evolutionary Computation Conference}{July 8--12, 2020}{Canc\'{u}n, Mexico}
\acmBooktitle{Genetic and Evolutionary Computation Conference (GECCO '20), July 8--12, 2020, Canc\'{u}n, Mexico}
\acmPrice{15.00}
\acmDOI{10.1145/3377930.3389820}
\acmISBN{978-1-4503-7128-5/20/07}

\begin{document}

\title[Analyzing Adaptive Parameter Landscapes in PAMs for DE]{Analyzing Adaptive Parameter Landscapes in Parameter Adaptation Methods for Differential Evolution}

%\title[Analyzing Adaptive Parameter Landscapes in DE]{Analyzing Adaptive Parameter Landscapes in\\Differential Evolution}

%\title[An Adaptive Parameter Landscape Analysis in Differential Evolution]{An Adaptive Parameter Landscape Analysis in\\Differential Evolution}

%% \title[PAL]{An Analysis of Parameter Adaptation Landscapes of Adaptive Differential Evolution}
%% \title[An Analysis of Parameter Adaptation Landscapes of Adaptive DE]{An Analysis of Parameter Adaptation Landscapes of Adaptive Differential Evolution}
%\title[PAL]{A Parameter Adaptation Landscape Analysis of Adaptive Differential Evolution}
%\title[PAL]{A Parameter Adaptation Landscape Analysis of Adaptive Differential Evolution}

%Automatic Algorithm Configuration
%Using an Unbounded External Archive}

%% \author{\hspace{1em}}
%% \affiliation{%
%%   \institution{\hspace{1em}}
%%   \streetaddress{\hspace{1em}}
%%   \city{\hspace{1em}\\\hspace{1em}} 
%% %  \state{\hspace{1em}} 
%%   \postcode{\hspace{1em}}
%% }

%% \author{\hspace{1em}}
%% \affiliation{%
%%   \institution{
%% %Faculty of Environment and Information Sciences,
%%     \hspace{1em}\\
%%     \hspace{1em}\\
%%   \hspace{1em}}
%% }
%% %\email{\hspace{1em}}

\author{Ryoji Tanabe}
\affiliation{%
  \institution{
%Faculty of Environment and Information Sciences,
    Yokohama National University\\
  Yokohama, Japan}
}
\email{rt.ryoji.tanabe@gmail.com}

\input{abstract.tex}
%% \begin{abstract}
%% This paper provides a sample of a \LaTeX\ document which conforms,
%% somewhat loosely, to the formatting guidelines for
%% ACM SIG Proceedings.\footnote{This is an abstract footnote}
%% \end{abstract}

\begin{CCSXML}
<ccs2012>
<concept>
<concept_id>10002950.10003714.10003716.10011136.10011797.10011799</concept_id>
<concept_desc>Mathematics of computing~Evolutionary algorithms</concept_desc>
<concept_significance>500</concept_significance>
</concept>
</ccs2012>
\end{CCSXML}

\ccsdesc[500]{Mathematics of computing~Evolutionary algorithms}

\keywords{DE, parameter adaptation methods, landscape analysis}

%\title[Analyzing Adaptive Parameter Landscapes in DE]{Analyzing Adaptive Parameter Landscapes in Parameter Adaptation Methods in Differential Evolution}

%real-coded crossover methods
\maketitle

\input{introduction.tex}

\input{preliminaries.tex}

\input{proposed_method.tex}

\input{setting.tex}

\input{results.tex}

\input{conclusion.tex}

%% \input{samplebody-conf}

\section*{Acknowledgments}

This work was supported by Leading Initiative for Excellent Young Researchers, MEXT, Japan.

\bibliographystyle{ACM-Reference-Format}
\bibliography{reference} 

\end{document}

%% file: abstract.tex
\begin{abstract}

  %A number of adaptive differential evolution (DE) algorithms have been proosed in the lirature.

%  Altough a number of adaptive evolutionary algorihms have been proosed in the lirature, their parameter adaptation mechanisms have not been well understtod.

%  Since the the scale factor $F$ and the crossover rate $\CR$ significanlty influence the performance of differential evolution (DE),

Since the scale factor and the crossover rate significantly influence the performance of differential evolution (DE), parameter adaptation methods (PAMs) for the two parameters have been well studied in the DE community.
Although PAMs can sufficiently improve the effectiveness of DE, PAMs are poorly understood (e.g., the working principle of PAMs).
One of the difficulties in understanding PAMs comes from the unclarity of the parameter space that consists of the scale factor and the crossover rate.
This paper addresses this issue by analyzing {\em adaptive parameter landscapes} in PAMs for DE.
First, we propose a concept of an adaptive parameter landscape, which captures a moment in a parameter adaptation process.
For each iteration, each individual in the population has its adaptive parameter landscape.
Second, we propose a method of analyzing adaptive parameter landscapes using a 1-step-lookahead greedy improvement metric.
Third, we examine adaptive parameter landscapes in three  PAMs by using the proposed method.
Results provide insightful information about PAMs in DE.

\end{abstract}

%% file: introduction.tex
\section{Introduction}
\label{sec:introduction}

This paper considers a black-box numerical optimization. % with bound constraints.
These problems involve finding a $d$-dimensional solution $\vector{x} = (x_1, ..., x_d)^{\top}$ that minimizes a given objective function $f: \mathbb{R}^d \rightarrow \mathbb{R}$, $\vector{x} \mapsto f(\vector{x})$.
Any explicit knowledge of $f$ is not given in black-box optimization.

Differential Evolution (DE) is a variant of evolutionary algorithms (EAs) mainly for black-box numerical optimization \cite{StornP97}.
The results in the annual IEEE CEC competitions have shown that DE is competitive with more complex optimizers despite its relative simplicity.
A number of previous studies have also demonstrated the effectiveness of DE in real-world applications \cite{DasS11,DasMS16}.

%,
%,LoboLM07,KarafotiasHE15

Main control parameters in the basic DE \cite{StornP97} include the population size $n$, the scaling factor $F$, and the crossover rate $\CR$.
From the late 1990s to the early 2000s, it had been believed that the performance of DE is robust with respect to the settings of $F$ and $\CR$ \cite{StornP97}.
However, some studies in the mid-2000s demonstrated that the performance of DE is sensitive to the settings of the control parameters \cite{GamperleMK02,ZielinskiWLK06,BrestGBMZ06}.
In general, the performance of EAs significantly depends on the characteristics of a given problem and the state of the search progress \cite{EibenHM99}.
Thus, a fixed parameter setting (e.g., $F=0.5$ and $\CR=0.9$) does not yield the best performance of an EA.
For these reasons, DE algorithms that automatically adjust the control parameters (mainly $F$ and $\CR$) have received much attention in the DE community since the mid-2000s.
Representative adaptive DE algorithms include jDE \cite{BrestGBMZ06}, SaDE \cite{QinHS09}, JADE \cite{ZhangS09}, EPSDE \cite{MallipeddiSPT11}, and SHADE \cite{TanabeF13}.
These adaptive DE algorithms have mechanisms to adaptively adjust the $F$ and $\CR$ parameters during the search process.
Parameter control methods in EAs can be classified into deterministic, adaptive, and self-adaptive control methods \cite{EibenHM99}.
Although some DE algorithms with deterministic and self-adaptive approaches have been proposed (e.g., \cite{WangCZ11,OmranSE05}), adaptive approaches have mainly been studied in the DE community \cite{TanabeF19}.

As in \cite{TanabeF17,TanabeF19}, this paper explicitly distinguishes ``an adaptive DE'' and ``a parameter adaptation method (PAM) in an adaptive DE''.
While ``an adaptive DE'' is a complex algorithm that consists of multiple components, ``a PAM'' is a single component only for adaptively adjusting $F$ and $\CR$ values.
As explained in \cite{TanabeF19}, ``L-SHADE'' \cite{TanabeF14CEC} is ``an adaptive DE'' that mainly consists of the following four components: (a) the current-to-$p$best/1 mutation strategy \cite{ZhangS09}, (b) the binomial crossover, (c) the ``PAM'' in SHADE \cite{TanabeF13}, and (d) the linear population size reduction strategy.
In this paper, we are interested in (c) the ``PAM'' in SHADE, rather than L-SHADE.
%In this paper, we denote (c) a ``PAM'' in an adaptive DE X as P-X (e.g., P-jDE and P-SHADE).

%we are interested in (c) P-SHADE, rather than SHADE itself.
%While there have been a number of performance comparisons among such complex DE algorithms, recent work \cite{ZielinskiWL08, DrozdikAAT15} pointed out that there are few comparative studies of PCMs \emph{in isolation}.

\begin{table*}[t]
\begin{center}
  \caption{\small Summary of the three landscape analysis.}
%{\scriptsize
%{\footnotesize
{\small
  \label{tab:summary_landscapes}
\scalebox{1}[1]{ 
\begin{tabular}{lll}
\toprule
Landscape analysis & Target space & Height\\ %& DC & NS\\
\midrule
Fitness landscape analysis & Solutions of a given problem & Fitness (or objective) values of solutions\\
Parameter landscape analysis & Static parameters in an EA (e.g., DE) & Expected performance of an EA with parameters\\
Adaptive parameter landscape analysis & Dynamic parameters adjusted by a PAM & 1-step-lookahead greedy improvement metric (G1) of parameters\\
\midrule
\end{tabular}
}
}
\end{center}
\end{table*}
%

%Dick10,

While most previous studies focused on ``adaptive DE algorithms'' (e.g., \cite{SeguraCSL14,ZamudaB15}), only a few previous studies tried to examine ``PAMs'' in DE.
Zielinski et al. investigated the performance of some PAMs for constrained optimization in an isolated manner \cite{ZielinskiWL08}.
Similar benchmarking studies for multi- and single-objective optimization were performed in \cite{DrozdikAAT15,TanabeF19}, respectively.
In  \cite{TanabeF16}, a lower bound on the performance of PAMs was analyzed by using an oracle-based method for approximating an optimal parameter adaptation process in DE.
A simulation framework for quantitatively evaluating the adaptation ability of PAMs was also proposed in \cite{TanabeF17}.

One of the difficulties in analyzing PAMs comes from the unclarity of the control parameter space that consists of $F$ and $\CR$.
For each iteration $t$, a PAM generates a parameter pair of $F^t_i$ and $\CR^t_i$ values $\vector{\theta}^t_i = (F^t_i, \CR^t_i)$ for the $i$-th individual $\vector{x}^t_i$ in the population, where $i \in \{1, ..., n\}$.
It is desirable that a trial vector $\vector{u}^t_i$ (a child or a new solution) generated with $\vector{\theta}^t_i$ is better than its parent $\vector{x}^t_i$ in terms of their objective values.
Generating a good $\vector{\theta}^t_i$ can be viewed as a two-dimensional numerical optimization problem.
The goal of this problem is to find the optimal parameter pair $\vector{\theta}^{*,t}_i \in \vector{\Theta}^t_i$ that minimizes the objective value of a trial vector $\vector{u}^t_i$, where $\vector{\Theta}^t_i \subseteq \mathbb{R}^2$ is a set of all feasible pairs of $F$ and $\CR$ values.
Although the properties of $\vector{\Theta}^t_i$ ({\bf {\em NOT the optimal parameter}}) can provide insightful information about PAMs, they have never been analyzed in the DE community and the evolutionary computation community.

%
%It is almost impossible to analyze $\vector{\Theta}^t_i$ in a straightforward manner.
%since $\vector{\Theta}^t_i$ is different depending on the individual, the state of the search progress, and a given problem.

%Altough the properties of $\vector{\Theta}^t_i$ can provide helpful clues for designing an efficient PAM, they are unknown in the DE community (and the evolutionary computation community).
%Thus, the properties of $\vector{\Theta}$ is unclear.
%Nevertheless, analyzing $\vector{\Theta}$ is attractive since

%For example, if $\vector{\Theta}$ has a 

%There are significant obstacles to analyzing $\vector{\Theta}$.

%% Sampling a good pair of $F$ and $\CR$ values can be viewed as a two-dimentional numerical optimization problem that involves finding the best pair of $F$ and $\CR$ values. % that involves finding a 
%% However, the control parameter space that consists of $F$ and $\CR$.

%In other words, the 
%try to 
%In this paper, we fill this gap between the necessity and the understandability of the properties of $\vector{\Theta}$ by analyzing adaptive parameter landscapes in DE.
%,

%This paper addresses this issue by analyzing adaptive parameter landscapes in PAMs for DE.

This paper tries to understand $\vector{\Theta}^t_i$  by analyzing its adaptive parameter landscape.
The term ``{\em adaptive parameter landscapes}'' is a new concept proposed in this paper inspired by recent work on {\em parameter landscapes} \cite{YuanOBS12,PushakH18,HarrisonOE19}.
As reviewed in \cite{PitzerA12,MalanE13,MunozSKH15}, {\em fitness landscapes} have been well studied in the evolutionary computation community.
In contrast, the field of parameter landscape analysis is relatively new.
A parameter landscape consists of feasible parameter values in an EA.
The ``height'' in parameter landscapes is the expected performance (or the utility) of an EA with control parameters on training problem instances \cite{EibenS11}.
An adaptive parameter landscape proposed in this paper can be viewed as a dynamic version of a parameter landscape influenced by a PAM.
For each iteration $t$, each individual $\vector{x}^t_i$ in the population has its adaptive parameter landscape that consists of $\vector{\Theta}^t_i$.
A PAM can be intuitively analyzed by investigating its adaptive parameter landscapes.
Table \ref{tab:summary_landscapes} summarizes differences in fitness landscapes, parameter landscapes, and adaptive parameter landscapes.
They are explained in Sections \ref{sec:fl}, \ref{sec:pl}, and \ref{sec:apl}, respectively.

Our contributions in this paper are at least threefold:

%%   In contrast to fitness landscapes (see Section \ref{sec:fl}) and parameter landscapes (see Section \ref{sec:pl}), the proposed adaptive parameter landscapes handle a sequential changes in the parameter adaptation process.
%% Differences among the three landscapes (fitness landscapes, parameter landscapes, and adaptive parameter landscapes) are discussed in Section \ref{sec:apl}.  

\begin{enumerate}
\item We propose a concept of adaptive parameter landscapes, which are landscapes of dynamic parameters adjusted by PAMs.
  This is the first study to address such dynamically changing parameter landscapes in the DE community and the evolutionary computation community.
\item We propose a method of analyzing adaptive parameter landscapes using a 1-step-lookahead greedy improvement metric.
\item We examine adaptive parameter landscapes in three representative PAMs on the 24 BBOB functions \cite{hansen2012fun} by using the proposed method.
  Results provide insightful information about PAMs.
  Our observations are summarized in Section \ref{sec:conclusion}.  
\end{enumerate}  

%%   We propose the method for analyzing adaptive parameter landscapes. To visualizes landscapes, we introduce a new evolvabitliy metric, called the 1-step greedy fitness improvement (G1) metric. 
%% \item We analyze the adaptive parameter landscapes in DE. This is the first study to discuss adaptive parameter landscapes in the field of evolutionary computation.
%% \item We provide insightful information about the parameter adaptation in DE. For example, the adaptive parameter landscapes can be... Such an information gives a general rule of thumb to design a new parameter adaptation method in DE.

%% This paper is organized as follows:
%% First, Section \ref{sec:de}  reviews DE, as well as representative parameter adaptation mechanisms for DE.
%% Then, Section \ref{sec:approximated_ideal} introduces our notion of an ``optimal parameter adaptation process'' and proposes the GAO model, which approximates this optimal parameter adaptation process. Section \ref{sec:approximated_ideal} also explains the relationship between GAO and previous work.
%% Section \ref{sec:experiment} presents an empirical evaluation of the GAO model.
%% We conclude with a discussion of our results and directions for future work in Section \ref{sec:conclusion}.

The rest of this paper is organized as follows.
Section \ref{sec:preliminaries} provides some preliminaries. %, including DE, fitness landscape analysis, and 
Section \ref{sec:proposed_method} explains the concept of adaptive parameter landscapes and the proposed analysis method.
Section \ref{sec:setting} describes the setting of our computational experiments.
Section \ref{sec:results} shows analysis results.
Section \ref{sec:conclusion} concludes this paper.

%Section \ref{sec:related_work} describes related studies.
% experimental results and their analysis.

%% file: preliminaries.tex
\section{Preliminaries}
\label{sec:preliminaries}

First, Section \ref{sec:de} explains the basic DE with a PAM.
Then, Section \ref{sec:pam} describes three PAMs in DE (the PAMs in jDE \cite{BrestGBMZ06}, JADE \cite{ZhangS09}, and SHADE \cite{TanabeF13}).
Finally, Sections \ref{sec:fl} and \ref{sec:pl} explain fitness landscape analysis and parameter landscape analysis, respectively.

%% Here, we explain the basic DE algorithm and parameter adaptation methos in the five adaptive DEs. We also reviwe the previous studies.
%% including DE, fitness landscape analysis, and 

%% In this paper, we suppose a black-box continuous minimization problem with a bound constraint:
%% %
%% \begin{align}
%%   %  \label{eqn}
%%   \text{minimize}_{\vector{x} \in \mathbb{S}} \:\, f (\vector{x}),\notag
%%   %% \text{minimize} \:\, &f (\vector{x}),\notag
%%   %% \text{subject to  } \, x^{\rm lower}_i  \geq x_i \geq x^{\rm upper}_i, i \in \{1, ..., n\},\notag    
%% \end{align}
%% %
%% where $\mathbb{S}$ is the $n$-dimensional solution space bounded by $\vector{x}^{\rm lower}$ and $\vector{x}^{\rm upper}$ such that $x^{\rm lower}_i  \geq x_i \geq x^{\rm upper}_i$ for each $i \in \{1, ..., n\}$.
%% The objective function $f: \mathbb{S} \rightarrow \mathbb{R}$ is to be minimized.
%% In the black-box scenario, $f$ is 

%% an objective function

%% that involve finding a $n$-dimensinal solution  $\vector{x} = (x_1, ..., x_n)^{\rm T} \in [x^{\rm lower}_]^n \subseteq \mathbb{R}^n$.

%%  that minimizes an objective function $f: \mathbb{S} \rightarrow \mathbb{R}$, where $D$ is the dimensionality of the problem, and $\mathbb{S}$ is the feasible region of the search space.

%% We suppose a multi-objective minimization problem in this
%% paper.
%% A continuous optimization problem with a simple symmetric boundary con-

\subsection{The basic DE with a PAM}
\label{sec:de}

%% Here, we explain the basic DE algorithm  \cite{StornP97}.

Algorithm \ref{alg:de} shows the overall procedure of the basic DE algorithm with a PAM.
Below, we explain DE {\em in an unusual manner} for a better understanding of the proposed G1 metric in Section \ref{sec:g1}.

%This unusual explanation about binomial crossover is good for a good understanding the proposed fitness improvement indicator.
At the beginning of the search $t=1$, the population $\vector{P}^{t} = \{ \vector{x}^t_{1}, ..., \vector{x}^t_{n} \}$ is initialized (line 1), where $n$ is the population size.
For each $i \in \{1, ..., n\}$, $\vector{x}^t_{i}$ is the $i$-th individual in the population $\vector{P}^{t}$.
Each individual represents a $d$-dimensional solution of a problem.
For each $j \in \{1, ..., d\}$, $x^t_{i,j}$ is the $j$-th element of $\vector{x}^t_{i}$.

After the initialization of $\vector{P}^t$, the following steps (lines 2--14) are repeatedly performed until a termination condition is satisfied.
For each $\vector{x}^t_{i}$, a parameter pair $\vector{\theta}^t_i = (F^t_i, \CR^t_i)$ is generated by a PAM (line 4).
The scale factor $F^t_i > 0$ controls the magnitude of the mutation.
The crossover rate $\CR^t_i \in [0,1]$ controls the number of elements inherited from $\vector{x}^t_i$ to a trial vector (child) $\vector{u}^t_i$.
When $\vector{\theta}^t_i$ is fixed for all individuals in the entire search process, Algorithm \ref{alg:de} becomes the classical DE without any PAM  \cite{StornP97}.

A set of parent indices $\vector{R} = \{r_1, r_2, ...\}$ are randomly selected from $\{1, ..., n\} \setminus \{i\}$ such that they differ from each other (line 5).
For each $\vector{x}^t_{i}$, a mutant vector $\vector{v}^t_{i}$ is generated by applying a differential mutation to $\vector{x}^t_{r_1}, \vector{x}^t_{r_2}, ...$ (line 6).
Although a number of mutation strategies have been proposed in the literature \cite{DasS11}, we consider the following two representative mutation strategies:
\begin{align}
\label{eqn:rand1}
\vector{v}^t_{i} &= \vector{x}^t_{r_1} + F^t_{i} \: (\vector{x}^t_{r_2} - \vector{x}^t_{r_3}),\\
\label{eqn:ctp1}
\vector{v}^t_{i} &= \vector{x}^t_{i}  +  F^t_{i} \: (\vector{x}^t_{p{\rm best}} - \vector{x}^t_{i}) +  F^t_{i} \: ( \vector{x}^t_{r_1} - \tilde{\vector{x}}^t_{r_2}),
\end{align}
where the strategy in \eqref{eqn:rand1} is rand/1 \cite{StornP97}, and the strategy in \eqref{eqn:ctp1} is current-to-$p$best/1 \cite{ZhangS09}.
The rand/1 strategy is the most basic strategy, and the current-to-$p$best/1 strategy is one of the most efficient strategies used in recent work (e.g., \cite{TanabeF13}, \cite{TanabeF14CEC}, \cite{ZhangS09}).
For each individual, the individual $\vector{x}^t_{p{\rm best}}$ is randomly selected from the top ``${\rm max}(\lfloor n \times p \rfloor , 2)$'' individuals in $\vector{P}^t$, where $p \in [0, 1]$ controls the greediness of current-to-$p$best/1.
The individual $\tilde{\vector{x}}^{t}_{r_2}$ in \eqref{eqn:ctp1} is randomly selected from a union of $\vector{P}^t$ and an external archive $\vector{A}^t$, where inferior parent individuals are preserved in $\vector{A}^t$ (how to update $\vector{A}^t$ is explained later).

After the mutant vector $\vector{v}^t_i$ has been generated for each $\vector{x}^t_i$, a trial vector $\vector{u}^t_i$ is generated by applying crossover to $\vector{x}^t_i$ and $\vector{v}^t_i$ (lines 7--9).
In this paper, we use binomial crossover \cite{StornP97}, which is the most representative crossover method in DE.
First, a $d$-dimensional vector $\vector{s} = (s_1, ..., s_d)^{\rm T}$ is generated (line 7), where each element in $\vector{s}$ is randomly selected from $[0,1]$.
%Each element in $\vector{s}$ is a random number generated from a uniform distribution in the range $[0,1]$.
An index $j_{\rm rand}$ is also randomly selected from $\{1, ..., d\}$ (line 8).
Then, for each $i \in \{1, ..., n\}$, the trial vector $\vector{u}^t_i$ is generated as follows (line 9):
\begin{align}
  \label{eqn:bin}
  u^t_{i,j} = \begin{cases}
    v^t_{i,j} & \:  {\rm if} \: s_j \leq \CR^t_i \:\: {\rm or} \:\: j = j_{\rm rand}\\
    x^t_{i,j} & \:  {\rm otherwise}
  \end{cases},
\end{align}
where the existence of $j_{\rm rand}$ ensures that at least one element is inherited from $\vector{v}^t_i$ even when $\CR^t_i = 0$.
%Unless $\vector{x}^i \neq \vector{v}^i$, 
%where 

After the trial vector $\vector{u}^t_i$ has been generated for each $\vector{x}^t_i$, the environmental selection is performed in a pair-wise manner (lines 10--12).
For each $i \in \{1, ..., n\}$, $\vector{x}^t_{i}$ is compared with $\vector{u}^t_{i}$.
The better one between $\vector{x}^t_{i}$ and $\vector{u}^t_{i}$ survives to the next iteration $t+1$.
%If $f(\vector{u}^t_{i}) \leq f(\vector{x}^t_{i})$,  $\vector{x}^t_{i}$ is replaced with $\vector{u}^t_{i}$.
%
The individuals that were worse than the trial vectors are preserved in the external archive $\vector{A}$ used in \eqref{eqn:ctp1}.
When the size of the archive exceeds a pre-defined size, randomly selected individuals are deleted to keep the archive size constant.
%Individuals in $\vector{A}$ are used for the current-to-$p$best/1 mutation strategy.
After the environmental selection, some internal parameters in a PAM are updated (line 13).

\def\HiLi{\leavevmode\rlap{\hbox to \hsize{\color{black!7}\leaders\hrule height .8\baselineskip depth .5ex\hfill}}}

\IncMargin{0.5em}
\begin{algorithm}[t]
%\scriptsize
\footnotesize
%\small
%\SetAlgoLined
\SetSideCommentRight
%\KwData{this text}
%\KwResult{how to write algorithm with \LaTeX2e }
%\tcp{\  Initialization phase}
$t \leftarrow 1$, initialize $\vector{P}^t =\{ \vector{x}^{t}_1, ..., \vector{x}^{t}_n\}$ randomly\;
%Set $F$ and $\CR$ values\;
%$F_{i,t} \leftarrow 0.5$, $\CR_{i,t} \leftarrow 0.9$, $i \in \{1, ..., N\}$\;
%
\While{$\textsf{\upshape{The termination criteria are not met}}$}{
\For{$i \in \{1, ..., n\}$}{
      Sample a parameter pair $\vector{\theta}^t_i = (F^t_i, \CR^t_i)$\;
      \HiLi  $\vector{R} \leftarrow$ A set of randomly selected indices from $\{1, ..., n\} \setminus \{i\}$\;
      %$r_1, r_2, ...$ from $\{1, ..., n\} \backslash \{i\}$ such that they differ from each other\;
      $\vector{v}^t_i \leftarrow {\rm mutation}(\vector{P}^t, \vector{R}, F^t_i)$\;
      \HiLi  $\vector{s} \leftarrow$ A randomly generated $d$-dimensional vector  $(s_1, ..., s_d)^{\top}$\;      
      \HiLi  $j_{\rm rand} \leftarrow$ A randomly selected number from $\{1, ..., d\}$\;
    $\vector{u}^{t}_i \leftarrow {\rm crossover}(\vector{x}^t_i, \vector{v}^t_i, C^t_i, \vector{s}, j_{\rm rand})$\;
    %    Generate the trial vector $\vector{u}^{i, t}$ by crossing $\vector{x}^{i, t}$ and $\vector{v}^{i, t}$ using a crossover method with $\CR$ (see Algorithms \ref{alg:bin-cross}, \ref{alg:exp-cross}, and \ref{alg:s-exp-cross})\;
  }
  \For{$i \in \{1, ..., n\}$}{
    \lIf{$f(\vector{u}^t_{i}) \leq f(\vector{x}^t_{i})$} {
      $\vector{x}^{t+1}_{i} \leftarrow \vector{u}^{t}_{i}$\;
    }
    \lElse{
      $\vector{x}^{t+1}_{i} \leftarrow \vector{x}^{t}_{i}$\;
    }    
  }
  Update internal parameters for adaptation of $F$ and $\CR$\;
  $t \leftarrow t+1$\;
}
\caption{The basic DE algorithm with a PAM}
\label{alg:de}
\end{algorithm}\DecMargin{0.5em}

\subsection{Three PAMs for DE}
\label{sec:pam}

We briefly explain the following three representative PAMs for DE: the PAM for jDE (P-jDE), the PAM for JADE (P-JADE), and the PAM for SHADE (P-SHADE).
%We re-emphasize that we are interested in PAMs in DE.
Our explanations are based on \cite{TanabeF19}.
Although we briefly explain the three PAMs due to space constraint, their detailed explanations with precisely described pseudo-codes can be found in \cite{TanabeF19}.
Below, the generation of the trial vector $\vector{u}^{t}_{i}$ is said to be {\em successful} if $f(\vector{u}^{t}_{i}) \leq f(\vector{x}^{t}_{i})$ (line 11 in Algorithm \ref{alg:de}).
Otherwise, the generation of $\vector{u}^{t}_{i}$ is said to be {\em failed}.

%% P-jDE assigns a pair of F i,t and C i,t to each individual x i,t , and
%% If the trial vector u i,t is better
%% than its parent individual x i,t , the newly generated parameter
%% is inherited by x i,t+1 .

P-jDE \cite{BrestGBMZ06} assigns a pair of $F^t_{i}$ and $\CR^t_{i}$ to each $\vector{x}^{t}_i$ in $\vector{P}^t$.
At the beginning of the search, these parameter values are initialized to $F^t_{i} = 0.5$ and $\CR^t_{i} = 0.9$ for each $i \in \{1, ..., n\}$.
%$F^t_{i}$ and $\CR^t_{i}$ are randomly regenerated with a predefined probability during the search process.
%
In each iteration $t$, $F^{{\rm trial},t}_{i}$ and $\CR^{{\rm trial},t}_{i}$ used for the generation of $\vector{u}^t_i$ are inherited from $\vector{x}^t_i$ as follows: $F^{{\rm trial}, t}_{i} = F^{t}_{i}$ and $\CR^{{\rm trial}, t}_{i} = \CR^{t}_{i}$.
However, with pre-defined probabilities $\tau_F$ and $\tau_{\CR}$, these values are randomly generated as follows: $F^{{\rm trial}, t}_{i} = {\rm randu}[0.1,1]$ and $\CR^{{\rm trial}, t}_{i} = {\rm randu}[0,1]$.
Here, ${\rm randu}[a, b]$ is a value selected uniformly randomly from $[a, b]$.
In general, the two hyper-parameters $\tau_F$ and $\tau_{\CR}$ are set to $0.1$.
%$\tau_F$ and $\tau_{\CR}$ are hyper-parameters of P-jDE.
%
When the generation of $\vector{u}^{i,t}$ is successful, $F^{t+1}_{i} = F^{{\rm trial}, t}_{i}$ and $\CR^{t+1}_{i} = \CR^{{\rm trial}, t}_{i}$.
Otherwise, $F^{t+1}_{i} = F^{t}_{i}$ and $\CR^{t+1}_{i} = \CR^{t}_{i}$.

%% When the generation of $\vector{u}^{i,t}$ is successful, $F^{t+1}_{i} = F^{{\rm trial}, t}_{i}$ and $\CR^{t+1}_{i} = \CR^{{\rm trial}, t}_{i}$.
%% Otherwise, $F^{t+1}_{i} = F^{t}_{i}$ and $\CR^{t+1}_{i} = \CR^{t}_{i}$.

%Each $\vector{u}^{i,t}$ is generated by using $F^{\rm trial}_{i,t}$ and $\CR^{\rm trial}_{i,t}$ in \eqref{eqn:jde-f} and \eqref{eqn:jde-cr}.

%% In each iteration $t$, $F^{{\rm trial},t}_{i}$ and $\CR^{{\rm trial},t}_{i}$ used to generate the trial vector $\vector{u}^t_i$ are generated as follows:
%% %
%% {\small
%% \begin{align}
%% \label{eqn:jde-f}
%% %\small
%% %\small
%% %\tiny
%% F^{{\rm trial}, t}_{i} &= \begin{cases}
%% {\rm randu}[0.1,1] &   {\rm if} \: {\rm randu}[0,1] < \tau_F\\
%% F^t_{i} &   {\rm otherwise}
%%   \end{cases},
%% \\
%% \label{eqn:jde-cr}
%% %\small
%% %\small
%% \CR^{{\rm trial},t}_{i} &= \begin{cases}
%% {\rm randu}[0,1]  &   {\rm if} \: {\rm randu}[0,1] < \tau_{\CR}\\
%% \CR^t_{i} &   {\rm otherwise}
%%   \end{cases},
%% \end{align}
%% }%

%for parameter adaptation ($\tau_F = \tau_{\CR} = 0.1$ is the recommended setting). 

%for mutation and crossover

%\subsubsection{P-JADE}

P-JADE \cite{ZhangS09} adaptively adjusts $F$ and $\CR$ values using two meta-parameters $\mu_{F}$ and  $\mu_{\CR}$, respectively.
For $t=1$, both $\mu_{F}$ and $\mu_{\CR}$ are initialized to $0.5$.
For each iteration $t$, $F^t_{i}$ and $\CR^t_{i}$ are generated as follows: $F^t_{i} = {\rm randc}(\mu_{F}, 0.1)$ and $\CR^t_{i} = {\rm randn}(\mu_{\CR}, 0.1)$.
Here, ${\rm randn}(\mu, \sigma^2)$ is a value selected randomly from a Normal distribution with mean $\mu$ and variance $\sigma^2$.
Also, ${\rm randc}(\mu, \sigma)$ is a value selected randomly from a Cauchy distribution with location parameter $\mu$ and scale parameter $\sigma$.
At the end of each iteration, $\mu_{F}$ and $\mu_{\CR}$ are updated based on sets $\vector{S}^F$ and $\vector{S}^{\CR}$ of successful $F$ and $C$ values as follows: $\mu_{F} = (1 - c) \: \mu_{F} + c \: {\rm mean}_L(\vector{S}^{F})$ and $  \mu_{\CR} = (1 - c) \: \mu_{\CR} + c \: {\rm mean}_A(\vector{S}^{\CR})$.
Here, $c \in [0,1]$ is a learning rate.
In general, $c=0.1$.
While ${\rm mean}_A(\vector{S}^F)$ is the arithmetic mean of $\vector{S}^F$, ${\rm mean}_L(\vector{S}^{\CR})$ is the Lehmer mean of $\vector{S}^{\CR}$.

%: ${\rm mean}_L(\vector{S}) = (\sum_{s \in \vector{S}} s^2) / (\sum_{s \in \vector{S}} s)$.

%% For each iteration $t$, successful $F$ and $\CR$ parameters are stored into sets $\vector{S}^{F}$ and $\vector{S}^{\CR}$, respectively. 
%% We use $\vector{S}$ to refer to $\vector{S}^{F}$ or $\vector{S}^{\CR}$ wherever the ambiguity is irrelevant or resolved by context.
%At the end of the iteration, $\mu_{F}$ and $\mu_{\CR}$ are updated as: 

%% When $F_{i,t}> 1$, $F_{i,t}$ is truncated to $1$. When $F_{i,t} \leq 0$, the new $F_{i,t}$ is repeatedly generated  in order to generate a valid value.
%% %In case a value for $\CR_{i,t}$ outside of $[0,1]$ is generated, it is replaced by the limit value (0 or 1) closest to the generated value.

%% %
%% %
%% \begin{align}  
%% \label{eqn:JADE_f_gen}
%% F_{i,t} &= {\rm randc}(\mu_{F}, 0.1)
%% %
%% %
%% \CR_{i,t} &= {\rm randn}(\mu_{\CR}, 0.1)
%% \end{align}
%
%% ${\rm randc}(\mu_{F}, \sigma)$ are values selected randomly from a Cauchy distribution with location parameter $\mu_{F}$ and scale parameter $\sigma$.
%% ${\rm randn}(\mu_{\CR}, \sigma^2)$ are values selected randomly from a normal distribution with mean $\mu_{\CR}$ and variance $\sigma^2$.
%% % 

%% %
%% %
%% \begin{align}  
%% \small
%% \label{eqn:update_mu_f}
%%   \mu_{F} &= (1 - c) \: \mu_{F} + c \: {\rm mean}_L(\vector{S}^{F})\\
%% \label{eqn:update_mu_cr}
%%   \mu_{\CR} &= (1 - c) \: \mu_{\CR} + c \: {\rm mean}_A(\vector{S}^{\CR})
%% \end{align}
%

%\subsubsection{P-SHADE}

P-SHADE \cite{TanabeF13} adaptively adjusts $F$ and $\CR$ using historical memories $\vector{M}^{F} = (M^{F}_1, ..., M^{F}_H)$ and $\vector{M}^{\CR} = (M^{\CR}_1, ..., M^{\CR}_H)$.
Here, $H$ is a memory size.
$H=10$ was recommended in \cite{TanabeF17}.
For $t=1$,  all elements in $\vector{M}^{F} $ and $ \vector{M}^{\CR}$ are initialized to $0.5$.
As reviewed in \cite{TanabeF19}, some slightly different versions of P-SHADE have been proposed by the same authors.
As in \cite{TanabeF19}, this paper considers the simplest version of P-SHADE presented in \cite{TanabeF17}.
In each iteration $t$,  $F^t_{i}$ and $\CR^t_{i}$ are generated as follows: $F^t_{i} = {\rm randc}(M^{F}_{r}, 0.1)$ and $\CR^t_{i} = {\rm randn}(M^{\CR}_{r}, 0.1)$, where $r$ is an index randomly selected from $\{1, ..., H\}$.
At the end of each iteration, the $k$-th elements in $\vector{M}^{F}$ and $\vector{M}^{\CR}$ are updated as follows: $M^{F}_k = {\rm mean}_L(\vector{S}^{F})$ and $M^{\CR}_k = {\rm mean}_L(\vector{S}^{\CR})$.
An index $k \in \{1, ..., H\}$ represents the position to be updated and is incremented on every update.
If $k > H$, $k$ is re-initialized to $1$.

%% At the beginning of the search, $k$ is initialized to $1$, 
%% and incremented whenever a new element is inserted into the history.
%% 
%If $F_{i,t}$ and $\CR_{i,t}$ are outside the range $[0, 1]$, they are adjusted/regenerated according to the procedure of  PCM-JADE.

%
%
%
%% \begin{align}
%% %\small
%% \label{eqn:SHADE_f_gen}
%% F_{i,t} &= {\rm randc}(M^{F}_{r_{i,t}}, 0.1)\\  
%% \label{eqn:SHADE_cr_gen}
%% \CR_{i,t} &= {\rm randn}(M^{\CR}_{r_{i,t}}, 0.1)
%% \end{align}
%

%
%% %
%% \begin{align}
%% %\footnotesize
%% \label{eqn:shade_update_f}
%% M^{F}_k &= {\rm mean}_L(\vector{S}^F)\\
%% \label{eqn:shade_update_cr}
%% M^{\CR}_k &= {\rm mean}_L(\vector{S}^{\CR})
%% \end{align}
%

\subsection{Fitness landscape}
\label{sec:fl}

%% %Altough the term ``the fitness landscape'' is loosely defined in general, its formal define is described in
%% (Pitzer and Affenzeller 2012).
%% Simply, fitness landscapes are defined as follows:

According to Pitzer and Affenzeller \cite{PitzerA12}, a fitness landscape $\mathcal{L}_f$ in a numerical optimization problem is defined by a 3-tuple as follows:
\begin{align}
  \label{eqn:fl}
%\vector{\theta}^* \in \argmin_{\vector{\theta} \in \vector{\Theta}} c(\vector{\theta})  
\mathcal{L}_{f} = (\mathbb{X}, f, D),
\end{align}
where $\mathbb{X} \subseteq \mathbb{R}^d$ is the solution space (i.e., a set of all feasible solutions $\vector{x}$).
Also, $f: \vector{x} \mapsto f(\vector{x})$ is the objective function of a given problem.
$D: \vector{x} \times \vector{x} \mapsto \mathbb{R}$ is a distance function between two solutions (e.g., the Euclidean distance).

%Useful information can be obtained by analyzing $\mathcal{L}_f$.
An analysis of $\mathcal{L}_f$ can provide useful information even for black-box optimization.
For example, if the features of a given problem (e.g., ruggedness and neutrality) becomes clear by analyzing $\mathcal{L}_f$, an appropriate optimizer can be selected \cite{MunozSKH15}.
A number of methods for analyzing $\mathcal{L}_f$ have been proposed in the literature \cite{PitzerA12,MalanE13}.
Representative methods include fitness distance correlation (FDC) \cite{JonesF95}, dispersion metric (DISP) \cite{LunacekW06}, and evolvability \cite{SmithHLO02}.
Recently, more sophisticated methods have been proposed, such as exploratory landscape analysis (ELA) \cite{MersmannBTPWR11} and local optima networks (LON) \cite{AdairOM19}.
These methods can quantify at least one feature about $\mathcal{L}_f$.
For example, the FDC value represents a global structure of $\mathcal{L}_f$ based on the correlation between the distance from solutions to the optimal solution and their objective values.

%information landscape \cite{BorensteinP05}.

%For example, the FDC value represents a global structure of $\mathcal{L}_f$ based on the correlation between objective values of solutions and the distance from each solution to the optimal solution.

%ndicators that map objective vectors to a real number.

\subsection{Parameter landscape}
\label{sec:pl}

Roughly speaking, a parameter tuning problem \cite{EibenS11} involves finding a tuple $\vector{\theta}$ of control parameters that optimizes the empirically estimated performance of an algorithm on a set of training problem instances.
For example, a parameter tuning problem for the basic DE with no PAM can be defined as a problem to find $\vector{\theta} = (F, \CR)$ that minimizes the average objective values of best-so-far solutions on the Sphere, Rastrigin, and Rosenbrock functions.
In general, the parameter tuning problem addresses only numerical parameters.

In contrast, an algorithm configuration problem \cite{HutterHLS09} addresses numerical, ordinal (e.g., low, medium, and high), and categorical parameters (e.g., the choice of mutation strategies).
According to \cite{Hoos12}, parameter tuning is a problem that involves only numerical parameters, while algorithm configuration is a problem that involves many categorical parameters.
The parameter tuning problem can be viewed as a special case of the algorithm configuration problem.

Parameter landscapes appear in parameter tuning problems.
Since it is difficult to define a distance function for categorical parameters, the field of parameter landscape analysis considers only numerical parameters as in \cite{PushakH18}.
The term ``parameter landscapes'' was first coined in \cite{YuanOBS12}.
Parameter landscapes were also denoted as ``performance landscapes'' \cite{YuanG07}, ``meta-fitness landscapes'' \cite{Pedersen10}, ``utility landscapes'' \cite{EibenS11}, ``ERT landscapes'' \cite{BelkhirDSS16}, ``parameter configuration landscapes'' \cite{HarrisonOE19}, and ``algorithm configuration landscapes'' \cite{PushakH18}.
To avoid any confusion, we use the term ``parameter landscapes'' throughout this paper.
Since only numerical parameters are considered, we believe that the term ``parameter landscapes'' is appropriate.
Although some previous studies (e.g., \cite{Bartz-BeielsteinLP10,LoshchilovSS12}) did not use the term ``landscapes'', they essentially investigated parameter landscapes.

According to Harrison et al. \cite{HarrisonOE19}, a parameter landscape $\mathcal{L}_{p}$ in a parameter tuning problem is formally defined as follows:
\begin{align}
  \label{eqn:pl}
\mathcal{L}_{p} = (\Theta, M, D),
\end{align}
where the definition of $\mathcal{L}_{p}$ in \eqref{eqn:pl} is a slightly different version of the original one in \cite{HarrisonOE19}.
$\Theta$ is the numerical parameter space (i.e., a set of all feasible parameters $\vector{\theta}$).
Also, $M: \vector{\theta} \mapsto M (\vector{\theta})$ is a performance metric that empirically estimates the performance of a given algorithm on a set of training problem instances (e.g., the average of objective values \cite{HarrisonOE19} and PAR10  \cite{PushakH18}).
Similar to \eqref{eqn:fl}, $D: \vector{\theta} \times \vector{\theta} \mapsto \mathbb{R}$ is a distance function between two parameters.

%Understanding Interactions among Genetic Algorithm Parameters \cite{DebA98}

%% Useful information can be obtained by analyzing the fitness landscape $\mathcal{L}_f$.
%% For example, if the features of $\mathcal{L}_f$ (e.g., ruggedness and neutrality) are clear, an appropriate optimizer can be selected.

Helpful information about parameter tuning and an algorithm can be obtained by analyzing $\mathcal{L}_{p}$.
For example, as mentioned in \cite{YuanG07},  if $\mathcal{L}_{p}$ is multimodal, a global parameter tuner may perform better than a local parameter tuner.
As demonstrated in \cite{YuanOBS12}, an influence of multiple parameters on the performance of an algorithm can be visually discussed by analyzing $\mathcal{L}_{p}$.

%% file: proposed_method.tex
\section{Proposed adaptive parameter landscape analysis}
\label{sec:proposed_method}

%This section explains the proposed method for analyzing adaptive parameter landscapes

First, Section \ref{sec:apl} explains the proposed concept of adaptive parameter landscapes.
Then, Section \ref{sec:g1} introduces a 1-step greedy fitness improvement (G1) metric, which is a performance metric for adaptive parameter landscapes.
Finally, Section \ref{sec:pal} proposes the method of analyzing adaptive parameter landscapes.

\subsection{Adaptive parameter landscape}
\label{sec:apl}

%As expained in Section \ref{sec:introduction}, since the control parameter space that consists of the scale factor and the crossover rate is unclear
%As expained in Section \ref{sec:introduction}, analyzing $\vector{\Theta}$ is attractive since helpful clues for designing an efficient PAM can be found from the properties of $\vector{\Theta}$.

We define an adaptive parameter landscape $\mathcal{L}_{a}$ in a PAM as follows:
\begin{align}
  \label{eqn:al}
\mathcal{L}_{a} = (\Theta^t_i, M, D),
\end{align}
where $\Theta^t_i$ is the numerical parameter space for the $i$-th individual in the population at iteration $t$ (i.e., a set of all feasible parameters $\vector{\theta}^t_i$).
The difference between $\mathcal{L}_{p}$ in \eqref{eqn:pl} and $\mathcal{L}_{a}$ in \eqref{eqn:al} is only the target space ($\Theta$ vs. $\Theta^t_i$).
While $\Theta$ in $\mathcal{L}_{p}$ is static, $\Theta^t_i$ in $\mathcal{L}_{a}$ is dynamic.
%Clearly, $\mathcal{L}_{a}$ is different depending on the individual and the state of the search process.
An adaptive parameter landscape can be viewed as a parameter landscape that captures a moment in a parameter adaptation process.

%captures a moment in a parameter adaptation process.

Our ultimate goal is to understand PAMs for DE.
While there have been significant contributions in analyzing DE itself in recent years (e.g., \cite{OparaA19,ArabasJ20}), only a few previous studies examined PAMs for DE (see Section \ref{sec:introduction}).
One reason is that very little is known about dynamically changing parameter spaces handled by PAMs.
We believe that this issue can be addressed by analyzing adaptive parameter landscapes.
A better understanding of adaptive parameter landscapes in PAMs can also lead to design a more efficient PAM.

Recall that Table \ref{tab:summary_landscapes} in Section \ref{sec:introduction} has already summarized the differences in a fitness landscape $\mathcal{L}_{f}$ in \eqref{eqn:fl}, a parameter landscape $\mathcal{L}_{p}$ in \eqref{eqn:pl}, and an adaptive parameter landscape $\mathcal{L}_{a}$ in \eqref{eqn:al}.
Very recently, Jankovic and Doerr \cite{JankovicD19} investigated dynamic fitness landscapes seen from CMA-ES \cite{HansenO01}.
They denoted their analysis as ``adaptive landscape analysis''.
While ``adaptive landscape analysis'' focuses on fitness landscapes of a problem, our adaptive parameter landscape analysis focuses on dynamic parameter landscapes adaptively adjusted by PAMs.
Thus, the names ``adaptive landscape analysis'' and ``adaptive parameter landscape analysis'' are similar, but they are totally different from each other.
%We emphasize that we are not interested in any fitness landscape.
As analyzed in \cite{DymondEH11,HarrisonEO18,BezerraLS18}, the best parameter settings in EAs depend on the maximum number of function evaluations when the performance of EAs is estimated based on final results (e.g., the objective value of the best-so-far solution at the end of each run).
We are interested in dynamically changing parameter landscapes, rather than such static parameter landscapes limited by a termination criterion.

Automated algorithm methods based on fitness landscape features of a given problem have been well studied in the evolutionary computation community \cite{MunozSKH15}.
Note that an adaptive parameter landscape analysis do not mean such a parameter selection approach that seeks the best static parameters (i.e., $F$ and $\CR$, not $F^t_i$ and $\CR^t_i$) based on fitness landscape features in a one-shot manner.
{\bf {\em For example, this paper is unrelated to \cite{BelkhirDSS16}.}}

\subsection{1-step-lookahead greedy improvement metric (G1)}
\label{sec:g1}

One critical obstacle in analyzing an adaptive parameter landscape $\mathcal{L}_{a}$ in \eqref{eqn:al} is how to define the performance metric $M$.
For an analysis of a parameter landscape $\mathcal{L}_{p}$ in \eqref{eqn:pl}, some performance metrics can be derived from the field of parameter tuning without any significant change (e.g., PAR10 as mentioned in Section \ref{sec:pl}).
In contrast, $M$ in $\mathcal{L}_{a}$ is not obvious.

%define the new evolvability indicator, called a 1-step greedy improvement indicator (G1).
%As explained in Section \ref{sec:de} with Algorithm \ref{alg:de}, the generation of the trial vector $\vector{u}^t_i$ and the environmental selection in DE is performed in a pair-wise manner.
%Thus, the selections in all $n$ individuals are performed independently from each other.
%The G1 metric uses this property of DE.

Here, we introduce a 1-step-lookahead greedy improvement (G1) metric as a performance metric $M$ for analyzing $\mathcal{L}_{a}$.
As explained in Section \ref{sec:de} using Algorithm \ref{alg:de}, DE generates the trial vector $\vector{u}^t_i$ for each parent individual $\vector{x}^t_i$ ($i \in \{1, ..., n\}$) at each iteration $t$.
A parameter pair of $F$ and $\CR$ values $\vector{\theta}^t_i = (F^t_i, \CR^t_i)$ is used for the generation of $\vector{u}^t_i$.
We define the G1 value of $\vector{\theta}^t_i$ as follows:
\begin{align}
  \label{eqn:g1}
  {\rm G1}(\vector{\theta}^t_i) = \begin{cases}
    \left| f (\vector{x}^t_i) - f (\vector{u}^t_i) \right| & \:  {\rm if} \: f (\vector{u}^t_i) < f (\vector{x}^t_i) \\
    0 & \:  {\rm otherwise}
  \end{cases}.
\end{align}
%

%Thus, $\vector{\theta}^t_i$ with a large G1 value can be considered as a good 

The ${\rm G1}(\vector{\theta}^t_i)$ value in \eqref{eqn:g1} represents how significantly $\vector{\theta}^t_i$ contributes to generate a better $\vector{u}^t_i$ than $\vector{x}^t_i$ in terms of the objective value.
A large G1 value indicates that the corresponding $\vector{\theta}^t_i$ can generate a good $\vector{u}^t_i$.
For example, let us consider the following three parameter pairs used for the generation of $\vector{u}^t_i$: $\vector{\theta}^t_{i,1}$, $\vector{\theta}^t_{i,2}$, and $\vector{\theta}^t_{i,3}$.
Their G1 values are also as follows: ${\rm G1}(\vector{\theta}^t_{i,1})= 0.7$, ${\rm G1}(\vector{\theta}^t_{i,2})= 1.2$, and ${\rm G1}(\vector{\theta}^t_{i,3})= 0$.
In this case, $\vector{\theta}^t_{i,2}$ is the best in the three parameter pairs in terms of G1.
The objective value of $\vector{x}^t_i$ can be significantly improved by using $\vector{\theta}^t_{i,2}$.
${\rm G1}(\vector{\theta}^t_{i,3})= 0$ means that $\vector{u}^t_i$ generated by using $\vector{\theta}^t_{i,3}$ is inferior (or equal) to $\vector{x}^t_i$.
Note that the G1 value is always non-negative.
In DE, an inferior trial vector compared to its parent individual cannot survive to the next iteration.
For this reason, we equally treat all parameter pairs of $F$ and $\CR$ values that generate worse trial vectors than their parent individuals.

The idea of measuring the fitness improvement value as in \eqref{eqn:g1} itself is not new at all.
Such an approach can be found in the literature (e.g., \cite{YangTY08,HansenAA15}).
In contrast to previous studies, the G1 metric aims to capture adaptive parameter landscapes by the proposed method explained in the next section.

%described in Section \ref{sec:pal}.

%Adaptive parameter landscapes based on the G1 metric can be viwed as the evolvability \cite{SmithHLO02} in the parameter space, not the solution space.

%we consider that any pair as equal.
%We are only interested in a pair of $F$ and $\CR$ values that improves the objective value of $\vector{x}^t_i$.
%Altough the negative G1 value can be defined by slightly revising \eqref{eqn:g1},  
%only a child with an inferior (or equallly) quality can be generated by using $\vector{\theta}_{i,3}$.

\subsection{Method for analyzing $\mathcal{L}_{a}$}
\label{sec:pal}

%% For each step of the
%% search (i.e., the parameter sampling of F i,t and C i,t for
%% the individual x i,t ), GAO randomly samples many possible
%% control parameter sets to retrospectively identify a control
%% parameter set which would have yielded the best-expected
%% result (with respect to 1-step-lookahead) on that step. By
%% repeating this process until the search terminates, GAO obtains
%% a parameter adaptation process that is approximately optimal
%% with respect to 1-step-lookahead.

%Our proposed method is inspired by the greedy approximate oracle method (GAO) \cite{TanabeF16}, which is a simulation based to approximate the optimal adaptation process of $F$ and $\CR$ in PAMs.

%We explain the proposed method\footnote{If this paper is accepted for publication, source code of the proposed method will be uploaded to the author's website.} of analyzing $\mathcal{L}_{a}$.

We explain the proposed method of analyzing $\mathcal{L}_{a}$.
Our proposed method can be incorporated into DE in Algorithm \ref{alg:de} with no change.
The procedure of our proposed method is totally independent from that of DE.
Thus, the search behavior of DE with and without our proposed method is exactly the same.

\begin{figure}[t]
   \centering
   \subfloat[$50 \times 50$ pairs]{\includegraphics[width=0.204\textwidth]{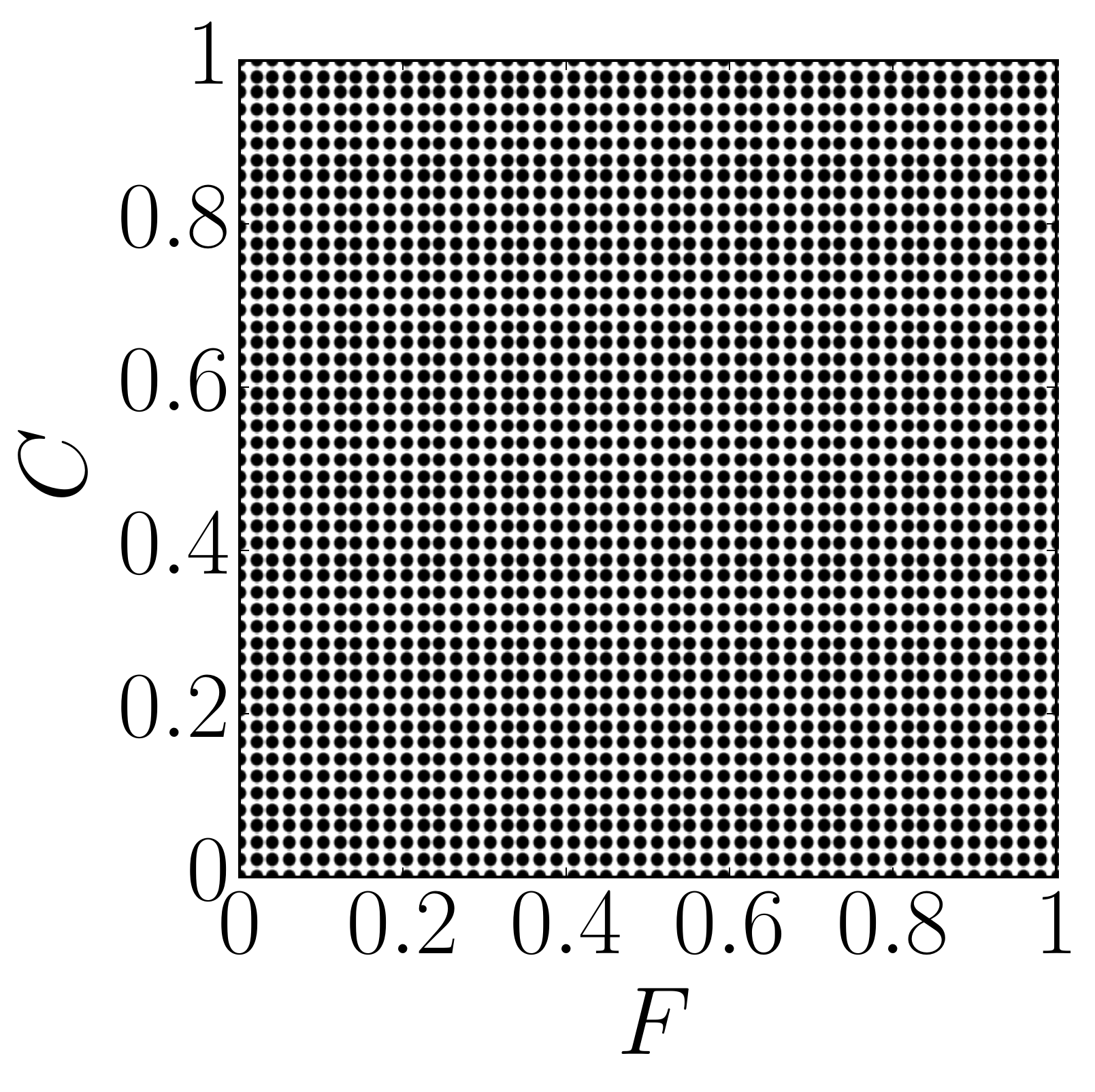}}
   \subfloat[Contour map of $\mathcal{L}_{a}$]{\includegraphics[width=0.24\textwidth]{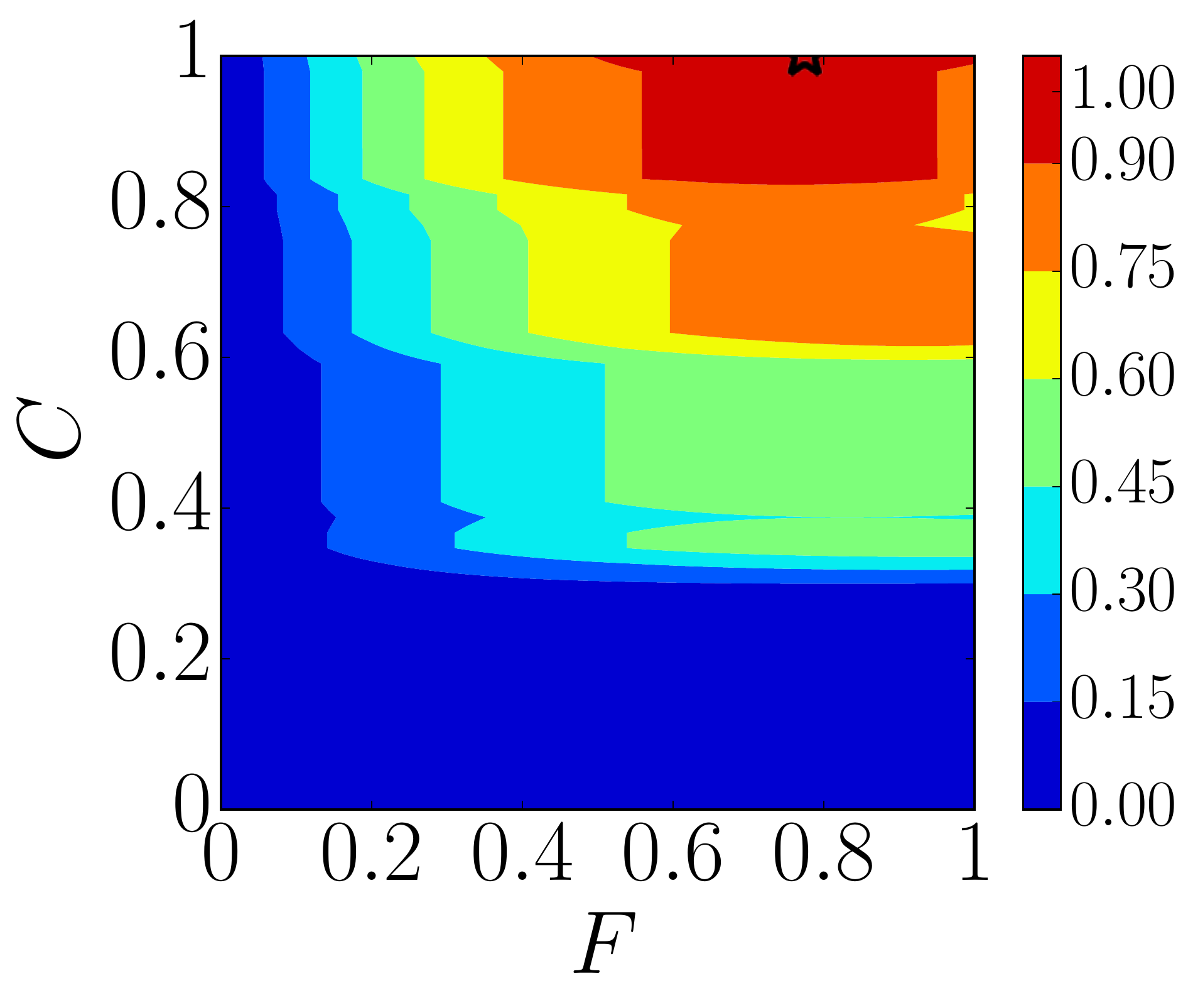}}
   \caption{
     (a) Distribution of $50 \times 50$ pairs of $F$ and $\CR$ values generated in a grid manner. (b) Contour map of its $\mathcal{L}_{a}$.
}
\label{fig:pal}
 \end{figure}

First, $m$ parameter pairs $\vector{\theta}^t_{i,1}, ..., \vector{\theta}^t_{i,m}$ are generated for each individual $\vector{x}^t_i$ at iteration $t$ (line 4 in Algorithm \ref{alg:de}).
Although any generation method can be used (e.g., the random sampling method), we generate $m$ parameter pairs in a grid manner in this study.
We generate $50 \times 50$ parameter pairs in the ranges $F \in [0,1]$ and $\CR \in [0,1]$.
Thus, $m=50 \times 50=2\,500$ in this study.
Figure \ref{fig:pal}(a) shows the distribution of the $50 \times 50$ parameter pairs.
We notice that any differential mutation strategy with $F=0$ does not work well, resulting poor performance of DE.
Just for the sake of simplicity, we include $F=0$ in the set of parameter pairs.
Since the G1 value of poor parameter pairs is 0, the inclusion $F=0$ does not significantly influence our analysis of $\mathcal{L}_{a}$.
We also notice that the upper value of $F$ is unbounded in principle, but it was generally set to $1$ in most previous studies (e.g., \cite{BrestGBMZ06,ZhangS09,MallipeddiSPT11,TanabeF13}).

Then, we calculate the G1 values of the $m$ parameter pairs ${\rm G1}(\vector{\theta}^t_{i,1}),$ $ ..., {\rm G1}(\vector{\theta}^t_{i,m})$ by simply generating $m$ trial vectors $\vector{u}^t_{i,1},$ $ ..., $ $\vector{u}^t_{i,m}$.
Their objective values $f(\vector{u}^t_{i,1}),$ $ ..., $ $f(\vector{u}^t_{i,m})$ are evaluated by $f$.
Then, for each $j \in \{1, ..., m\}$, ${\rm G1}(\vector{\theta}^t_{i,j})$ is calculated by \eqref{eqn:g1}.
It should be noted that $m$ extra function evaluations by $f$ are needed to calculate the $m$ objective values $f(\vector{u}^t_{i,1}),$ $ ..., $ $f(\vector{u}^t_{i,m})$.

In the proposed method, the $m$ extra function evaluations for each individual {\em are not counted} in the function evaluations used in the search.
This manner is similar to GAO \cite{TanabeF16}.
%We emphasize that we aim to analyze adaptive parameter landscapes, not to develop a new PAM.
The $m$ trial vectors are used only for adaptive parameter landscape analysis and are not used for the actual search.
Independently of the generation of the $m$ trial vectors, each individual $\vector{x}^t_i$ generates its trial $\vector{x}^t_i$ as in the traditional DE  (lines 4--9 in Algorithm \ref{alg:de}).
As mentioned above, the behavior of DE with any PAM does not change even when generating the $m$ extra trial vectors.

%not to develop a new PAM.
%Note that the calculation of the G1 value is fully independent from the main procedure of DE.
%The generated $n \times n$ parameter pairs are not used for the search in DE.
%% Note that we are interested in the G1 measure.

The stochastic nature of the basic DE is due to (1) the random selection of individual indices $\vector{R} = \{r_1, r_2, ...\}$ for mutation (line 5 in Algorithm \ref{alg:de}) and (2) the generation of random numbers $\vector{s} = (s_1, ..., s_d)^{\top}$ and $j_{\rm rand}$ for crossover  (lines 7--8 in Algorithm \ref{alg:de}).
Thus, the stochastic nature of DE can be ``virtually'' suppressed by fixing these random factors ($\vector{R}$, $\vector{s}$, and $j_{\rm rand}$).
For each iteration $t$, each individual $\vector{x}^t_i$ generates one actual trial vector $\vector{u}^t_{i}$ and the $m$ extra trial vectors $\vector{u}^t_{i,1},$ $ ..., $ $\vector{u}^t_{i,m}$ using the same $\vector{R}$, $\vector{s}$, and $j_{\rm rand}$.
When using the current-to-$p$best/1 strategy in \eqref{eqn:ctp1}, $\vector{x}^{t}_{p{\rm best}}$ must also be fixed.
Note that this suppression mechanism is used only to generate the $m$ trial vectors for each individual $\vector{x}^t_i$.

%% We calculate the G1 value of the $n$ parameter pairs ${\rm G1}(\vector{\theta}_{i,1}), ..., {\rm G1}(\vector{\theta}_{i,n})$ with the fixed random factor.
%As a result, $n$ trial vectors $\vector{u}_1, ..., \vector{u}_n$ are generated.

Figure \ref{fig:pal}(b) shows the contour map of $\mathcal{L}_{a}$ based on the $50 \times 50$ parameter pairs in Figure \ref{fig:pal}(a).
The height in $\mathcal{L}_{a}$ is the normalized G1 value.
For the sake of clarity, for each individual $\vector{x}^t_i$, we normalize all G1 values into the range $[0,1]$ by using the maximum G1 value ${\rm G1}^{\rm max}$ and the minimum G1 value ${\rm G1}^{\rm min}$ as follows: ${\rm G1}(\vector{\theta}^t_{i,j}) = ({\rm G1}(\vector{\theta}^t_{i,j}) - {\rm G1}^{\rm min}) / ({\rm G1}^{\rm max} - {\rm G1}^{\rm min})$, where $j \in \{1, ..., m\}$.
The contour map in Figure \ref{fig:pal}(b) is $\mathcal{L}_{a}$ in the 100-th individual in P-SHADE on the 20-dimensional $f_1$ in the BBOB function set \cite{hansen2012fun}.
Figure \ref{fig:pal}(b) shows $\mathcal{L}_{a}$ at the 100-th function evaluations.
We used the same experimental setting explained in Section \ref{sec:setting}.
%For details, see Section \ref{sec:setting}.
Details of the setting are described in Section \ref{sec:setting} later.
Figure \ref{fig:pal}(b) is the same with the most bottom left of Figure \ref{fig:c_sphere}.
In Figure \ref{fig:pal}(b), the parameter pair of $F=0.78$ and $\CR = 1$ is the best in the $m=50 \times 50$ parameter pairs in terms of G1.
As seen from Figure \ref{fig:pal}(b), the closer the parameter pair is to the best parameter pair, the better the G1 value.

%% file: setting.tex
\section{Experimental setup}
\label{sec:setting}

We performed all experiments using the COCO software (\url{https://github.com/numbbo/coco}), which is standard in the GECCO black-box optimization benchmarking (BBOB) workshops since 2009.
We used the 24 BBOB noiseless functions $f_1, ...,  f_{24}$ \cite{hansen2012fun}, which are grouped into the following five categories: separable functions ($f_1, ...,  f_5$),  functions with low or moderate conditioning ($f_6, ..., f_9$), functions with high conditioning and unimodal ($f_{10}, ..., f_{14}$), multimodal functions with adequate global structure ($f_{15}, ..., f_{19}$), and multimodal functions with weak global structure ($f_{20}, ...,  f_{24}$).
The dimensionality $d$ of the BBOB functions was set to $2, 3, 5, 10, 20$, and $40$.
For each problem instance, 15 runs were performed.
These settings strictly adhere to the standard benchmarking procedure in the GECCO BBOB workshops.
The maximum number of function evaluations was set to $10\,000 \times d$.

%We set the default hyperparametrs of the five parameter adaptation mechanisms reccomned by the corresponding articles as follows: $H = 10$ for P-SHADE, $\tau_F = 0.1$ and $\tau_{\CR} = 0.1$ for P-jDE.

We analyze the three PAMs (P-jDE, P-JADE, and P-SHADE) explained in Section \ref{sec:pam}.
Source code used in this study can be downloaded from \url{https://github.com/ryojitanabe/APL}.
We set their hyper-parameters to the values recommended by the corresponding articles.
As in \cite{BrestGBMZ06,ZhangS09,TanabeF13}, we set the population size $n$ to 100.
%We also investigate the performance of the standard DE with no parameter adaptation mechanism \cite{StornP97}.
%For the standard DE, we set $F$ and $\CR$ parameters to $0.5$ and $0.9$, respectively.
%These parameter settings are commonly used in previous work (e.g., \cite{BrestGBMZ06,ZhangS09}).
%
We used the rand/1 and current-to-$p$best/1 strategies described in Section \ref{sec:de}.
However, we show only results with current-to-$p$best/1 due to space constraints.
As in \cite{ZhangS09}, the control parameters of the current-to-$p$best/1 strategy were set as follows: $p = 0.05$ and $|\vector{A}| = n$.
We used binomial crossover.
%
%% Although restarts are mandatory for black-box optimization, we did not incorporate any restart strategy into DE.
%% This is because restarts can break the parameter adaptation process, making analysis difficult.

%we want to analyze the behavior of PAMs until the search stagnates.
%We want to emphasize that we are interested in how the parameter adaptation mechanisms in DE adaptively adjust $F$ and $\CR$ parameters, rather than which method performs the best.

%% Below, the DE with $F=0.5$ and $\CR=0.9$ is denoted as ``$F05\CR09$''. % or "No-PCM''?
%% Following previous work \cite{PosikK12a}, we set the population size $N$ to $5 \times D$ for  $D\geq 5$, and 20 for  $D\leq 3$.

%% file: results.tex
\newtheorem{observation}{Observation} % to emphasize key empirical observations

\begin{figure}[t]
   \centering
      \includegraphics[width=0.48\textwidth]{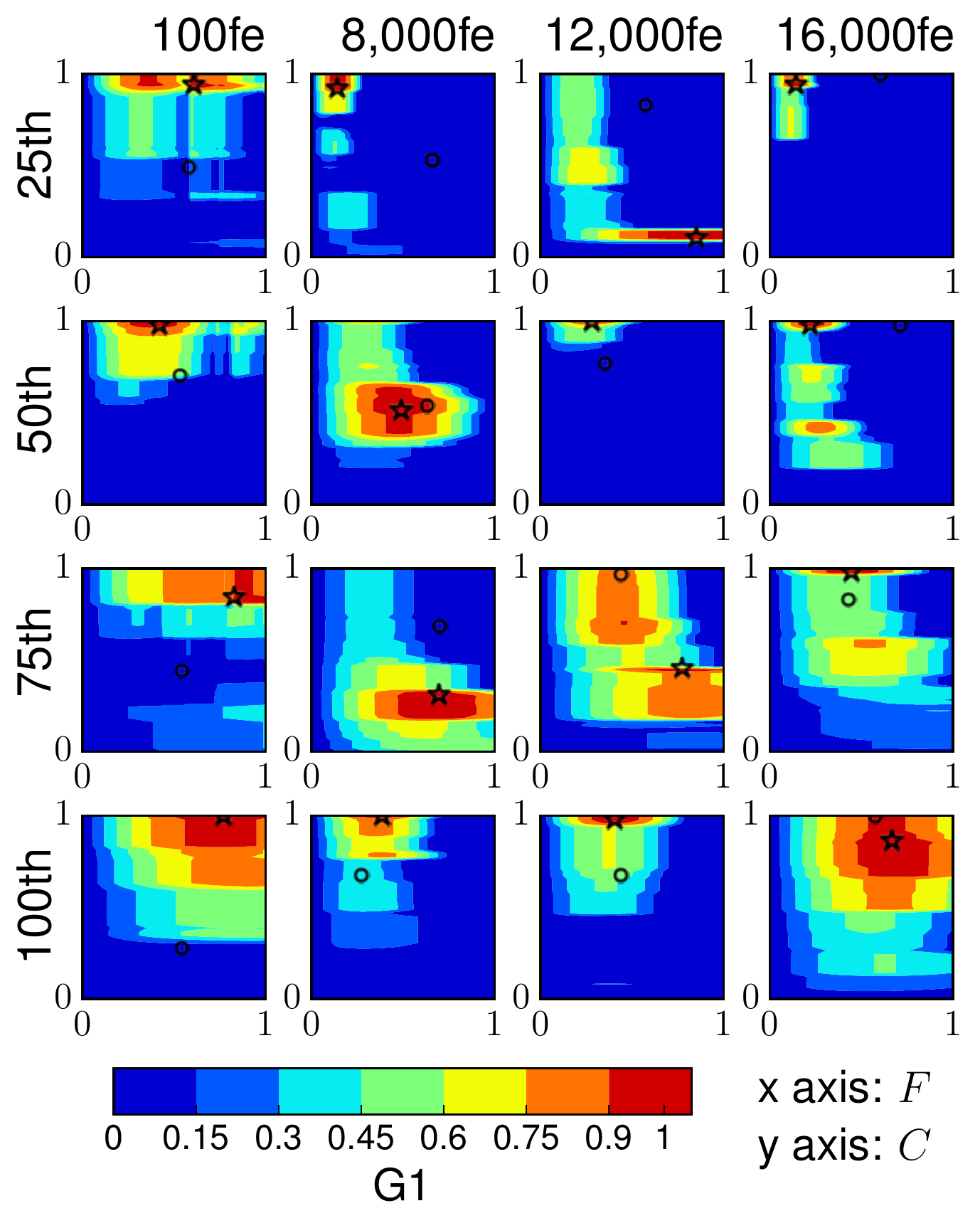}
   \\
   \caption{
     Contour maps of adaptive parameter landscapes in P-SHADE on $f_1$ with $d=20$.
   }
   \label{fig:c_sphere}
\end{figure}

\begin{figure}[t]
  \centering
\includegraphics[width=0.48\textwidth]{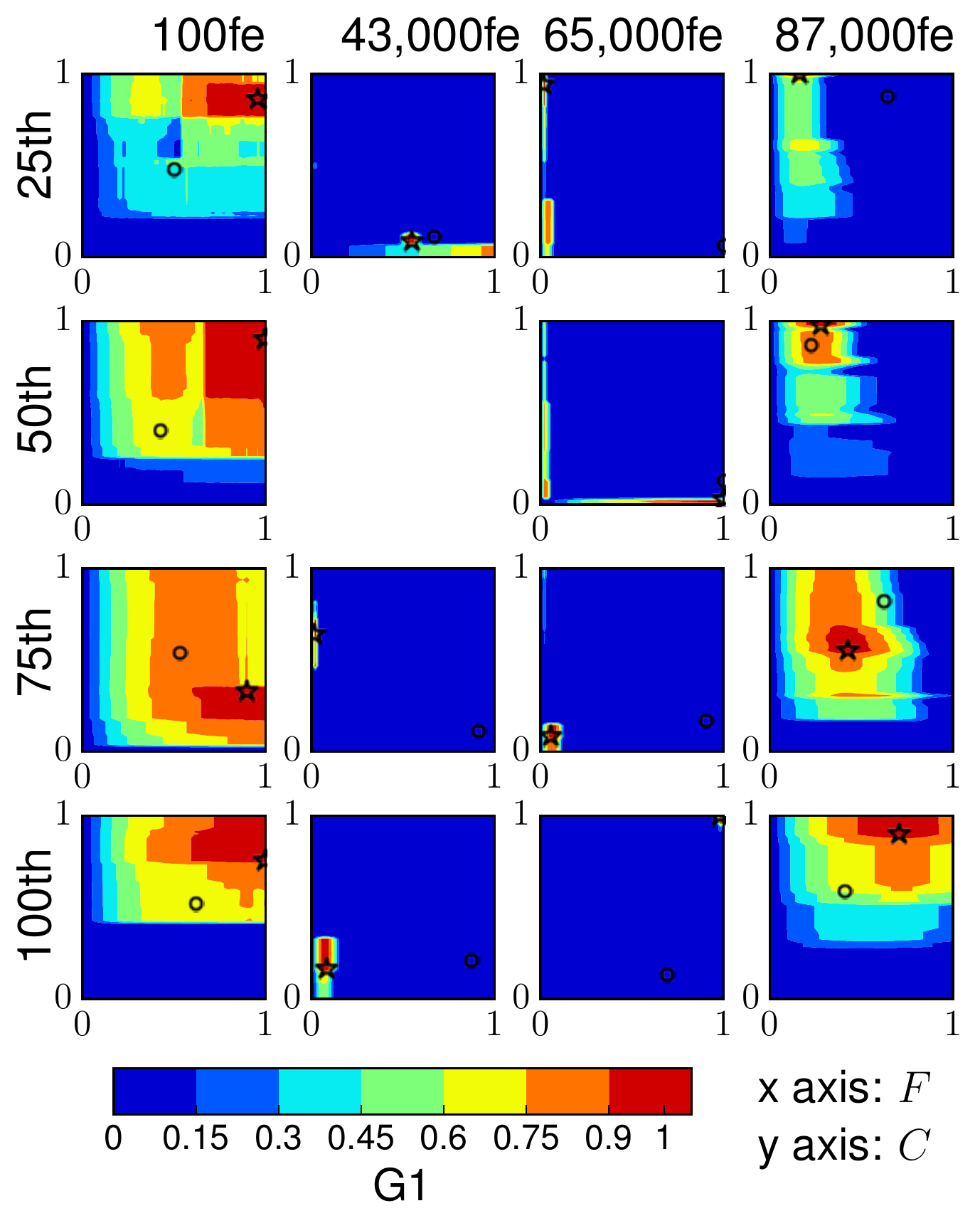}         
   \\
   \caption{
     Contour maps of adaptive parameter landscapes in P-SHADE on $f_3$ with $d=20$. %{\em The missing figure is not an error}. See Section \ref{sec:contour}.
   }
   \label{fig:c_rastrigin}
\end{figure}

%% \begin{figure*}[t]
%%   \centering
%% \includegraphics[width=0.9\textwidth]{graph/contour/shade_n100_current_to_pbest_1_bin_f8_d20.pdf}         
%%    \\
%%    \caption{
%%      Contour maps of adaptive parameter landscapes in P-SHADE with current-to-$p$best/1 on $f_8$ with $d=20$.   
%%    }
%%    \label{fig:c_rastrigin}
%% \end{figure*}

\section{Results}
\label{sec:results}

This section analyzes adaptive parameter landscapes in PAMs for DE by the proposed method.
Our findings are summarized in Section \ref{sec:conclusion}.
Section \ref{sec:contour} discusses the shapes of adaptive parameter landscapes by using contour maps as in Figure \ref{fig:pal}(b).
Section \ref{sec:fdc} examines adaptive parameter landscapes using landscape measures.

%% Subsection VI-B examines the nine quality indicators using the ranking information. Subsection
%% VI-C analyzes the generality of results for m = 3 with respect
%% to m. Subsection VI-D investigates a unary version of an M -
%% nary quality indicator.

\subsection{Analysis with contour maps}
\label{sec:contour}

% (i.e., $100$ iterations) 

Let us consider that a DE with the population size $n=100$ terminates the search at $10\,000$ function evaluations on a problem.
In this case, we can obtain $9\,900$ contour maps, where the first $100$ evaluations out of $10\,000$ are for the initialization of the population.
Also, we performed 15 runs of the 3 PAMs on the 24 BBOB functions with the 6 dimensionalities $d \in \{2, 3, 5, 10, 20, 40\}$.
Even if all runs terminate at $10\,000$ function evaluations, we can obtain $64\,152\,000$ contour maps $(=9\,900 \times 15 \times 3 \times 24 \times 6)$ contour maps.
It is impossible and meaningless to show $64\,152\,000$ contour maps in this paper.
%In addition, GECCO allows the authors to upload only a 10MB supplementary file.

%Since too many data do not 
%We sort results of 15 runs by the objective values.

For the above-mentioned reason, we ``thinned'' data as follows so that we can focus only on meaningful results.

\noindent {\bf $\bullet$} {\em Data of all runs.} For each PAM, we show results of a single run with a median best-so-far error value, which is the gap between the objective values of the best-so-far solution and the optimal solution.
When the error value is smaller than $10^{-8}$, it is treated as 0.
Ties are broken by the number of function evaluations that is used to find the best-so-far solution.

\noindent {\bf $\bullet$} {\em Data of all individuals.} For each iteration, first, all individuals are sorted based on their objective values in descending order.
Then, we show only adaptive parameter landscapes of the 25th, 50th, 75th, and 100th individuals out of 100 individuals.
Since a parameter pair for the best (1st) individual is seldom successful, we omit its results.
The reason is discussed in Section \ref{sec:fdc} later.

%Since the best (1st) individual seldom generates a better trial vector, we omit its results.

\noindent {\bf $\bullet$} {\em Data of all function evaluations.}
In order to reduce the computational cost of the proposed method, we calculate adaptive parameter landscapes only in every $1\,000$ function evaluations.
Also, we show results at $100$, $\lfloor 0.5 \, {\rm fe}^{\rm stop} \rfloor$, $\lfloor 0.75 \, {\rm fe}^{\rm stop} \rfloor$, and $\lfloor 1 \, {\rm fe}^{\rm stop} \rfloor$ function evaluations, where ${\rm fe}^{\rm stop}$ is the number of function evaluations when the best-so-far solution is updated last time.    

%For this reason, we do not show the parameter landscape of the 1th individual.
%We show only the results on $f_1$ (Sphere), $f_3$ (Rastrigin), $f_8$ (Rosenbrock) with $d=20$ in this paper.
%Thus, we show othere 

%These figures show results of the 25th, 50th, 75th, and 100th individuals at 100, $\lfloor 0.25 \, {FE}^{\rm stop} \rfloor$, $\lfloor 0.5 \, {FE}^{\rm stop} \rfloor$, $\lfloor 0.75 \, {FE}^{\rm stop} \rfloor$, and $\lfloor 1 \, {FE}^{\rm stop} \rfloor$ function evaluations, where ${FE}^{\rm stop}$ is the number of function evaluations when the best-so-far solution is updated last time.
%% These figures show results of a single run with a median performance among 15 runs.

Figures \ref{fig:c_sphere} and \ref{fig:c_rastrigin} show the contour maps of adaptive parameter landscapes in P-SHADE on $f_1$ and $f_3$ with $d=20$, respectively.
Here, $f_1$ and $f_3$ are modified versions of the Sphere function and the Rastrigin function, respectively.
In Figures \ref{fig:c_sphere} and \ref{fig:c_rastrigin}, ``fe'' stands for ``function evaluations''.
The x and y axes represent $F$ and $\CR$, respectively.
The star in each figure is the best parameter pair that maximizes the G1 value.
The circle in each figure is the parameter pair generated by the PAM.
See Section \ref{sec:pal} for how to generate Figures \ref{fig:c_sphere} and \ref{fig:c_rastrigin}.
When the G1 values of all $50 \times 50$ parameter pairs are 0 (i.e., no parameter pair can improve the individual), the adaptive parameter landscape is flat.
In such a case, we do not show results (e.g., the result of the 50th individual at $43\,000$ function evaluations in Figure \ref{fig:c_rastrigin}).
Figures S.73 in the supplementary file show error values of P-jDE, P-JADE, and P-SHADE on all 24 BBOB functions with $d=20$.
Note that we are not interested in benchmarking DE algorithms.
As shown in Figures S.73(a) and (c), P-SHADE found the optimal solution on $f_1$ and $f_3$ at about $16\,000$ and $87\,000$ function evaluations in a median run.
Due to space constraints, we show results of P-jDE, P-JADE, and P-SHADE on the 24 BBOB functions ($f_1$, ..., $f_{24}$) with $d=20$ in Figures S.1--S.72 in the supplementary file.
Although we show only the results of P-SHADE in this section, the qualitative results of the three PAMs are similar.
The results on the functions with $d \geq 5$ are also similar to those with $d=20$.

%% For each figure in Figures \ref{fig:c_sphere} and \ref{fig:c_sphere}, the horizontal axis represents $F$ values ($F \in [0, 1]$), and the vertical axis represents $\CR$ values ($\CR \in [0, 1]$).
%% We only show the parameter landscape of 25th, 50th, 75th, and 100th individuals at 100, and 1000, 3000, 4000, function evaluations.
%% Since the 1st individual does not often generate a better child, its adaptive parameter landscape is a flat in most  cases.
%% Details are discussed in Section \ref{} later.
%% For this reason, we do not show the results of the 1st individual.

Below, we discuss the shape of adaptive parameter landscapes obtained in this study.
Readers who want to quickly know our  observations can refer to Section \ref{sec:conclusion}.
As shown in Figures \ref{fig:c_sphere} and \ref{fig:c_rastrigin}, the shape of adaptive parameter landscapes is different depending on the search progress.
Since improving randomly initialized individuals is easy, the area with non-zero G1 values is large at the beginning of the search in most cases.
This means that generating a successful parameter pair of $F$ and $\CR$ is easy for PAMs in an early stage of evolution.
However, the area with non-zero G1 values decreases as the search progresses.
Thus, it is relatively difficult to generate a parameter pair of $F$ and $\CR$ that improves each individual in a mid stage of evolution.
As seen from Figures \ref{fig:c_sphere} and \ref{fig:c_rastrigin}, the shape of adaptive parameter landscapes is also different depending on the rank of each individual.
By comparing Figures \ref{fig:c_sphere} and \ref{fig:c_rastrigin}, we can see that generating a successful parameter pair on a multimodal function is more difficult than that on a unimodal function.
Adaptive parameter landscapes at $43\,000$ and $65\,000$ function evaluations in Figure \ref{fig:c_rastrigin} indicate that the area with non-zero G1 values is very small like needle-in-haystack landscapes.
In addition to the multimodality, the nonseparability is an important factor to determine the shape of adaptive parameter landscapes as seen from results on nonseparable unimodal functions ($f_6$--$f_{14}$) shown in Figures S.6--S.62 in the supplementary file.
Interestingly, as shown in adaptive parameter landscapes at $87\,000$ function evaluations in Figure \ref{fig:c_rastrigin}, the area with non-zero G1 values becomes large again in a late stage of evolution.
This is because the population has well converged to the optimal solution, and generating better trial vectors is not so difficult at such a situation on $f_3$.

\begin{figure}[t]
   \centering
   \subfloat[$f_{22}$]{\includegraphics[width=0.24\textwidth]{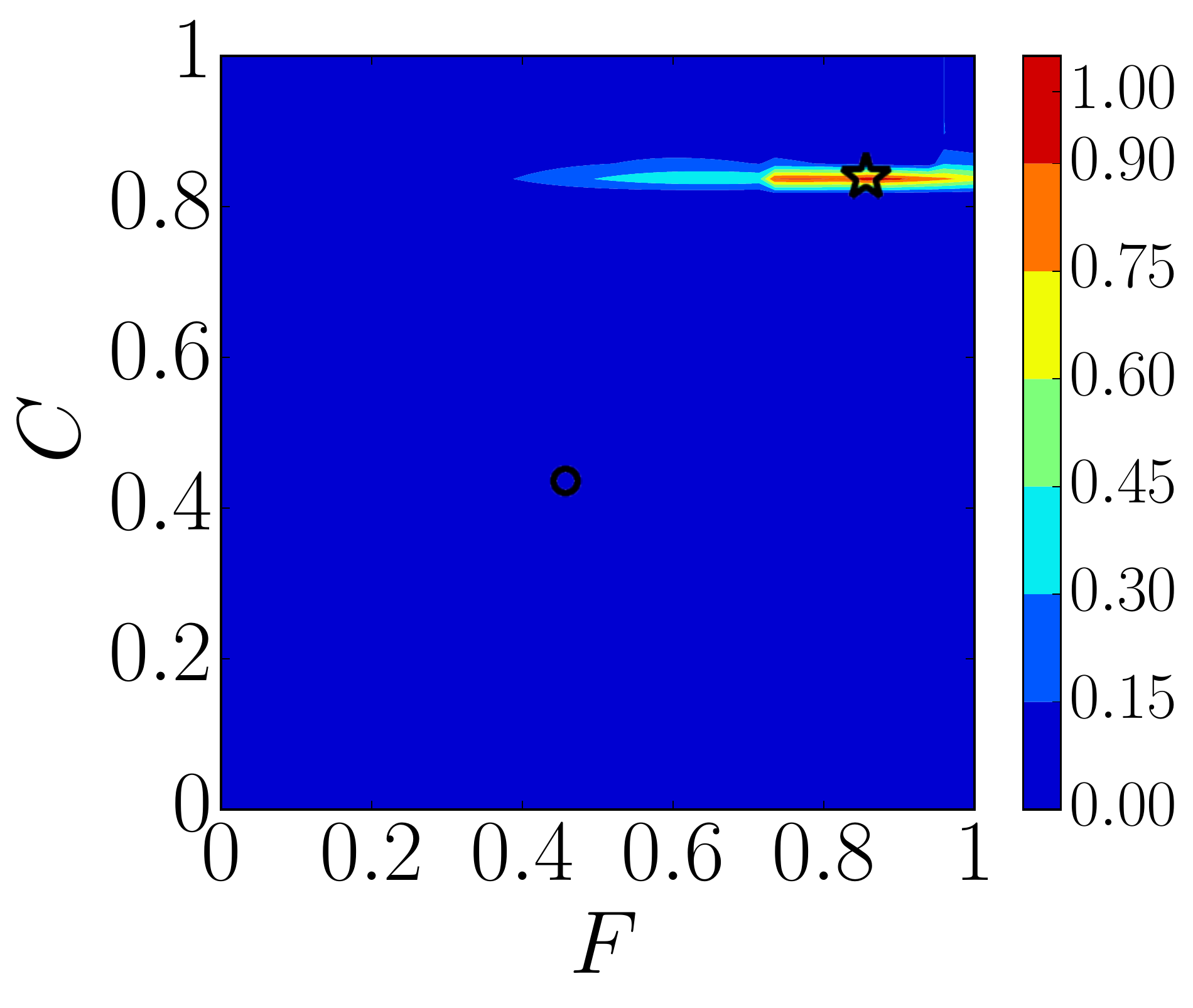}}
   \subfloat[$f_{23}$]{\includegraphics[width=0.24\textwidth]{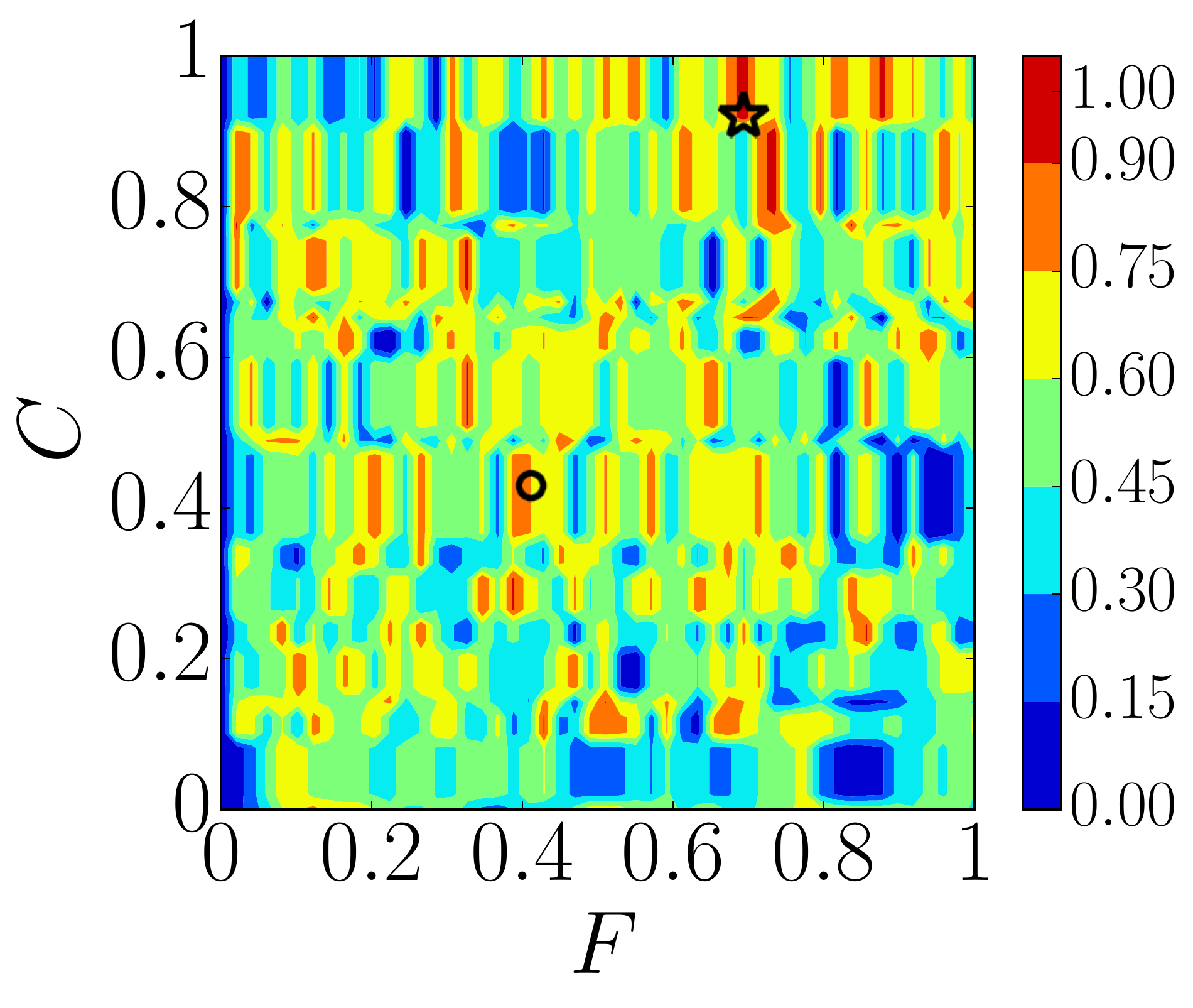}}         
%\subfloat[$89\,000$fe]{\includegraphics[width=0.24\textwidth]{graph/contour_singleplot/shade_n100_current_to_pbest_1_bin_f23_d20_89000fe.pdf}}
   \caption{
     Contour maps of adaptive parameter landscapes in P-SHADE on $f_{22}$ and $f_{23}$ with $d=20$.
   }
   \label{fig:cpam_f23}
\end{figure}

The shape of adaptive parameter landscapes is significantly influenced by the global structures of fitness landscapes.
For example, Figure \ref{fig:cpam_f23} shows that the contour maps of adaptive parameter landscapes in the 100-th individual of P-SHADE on $f_{22}$ and $f_{23}$ with $d=20$ at 100 function evaluations.
Figures \ref{fig:cpam_f23}(a) and (b) are parts of Figures  S.70 and S.71 in the supplementary file, respectively.
The original functions of $f_{22}$ and $f_{23}$ are the Gallagher's Gaussian 21 peaks function and the Katsuura function, which have fitness landscapes without any global structure.
Figure \ref{fig:cpam_f23}(a) shows that the adaptive parameter landscape on $f_{22}$ has only the small area with non-zero G1 values even at the beginning of the search.
Figure \ref{fig:cpam_f23}(b) also shows that the adaptive parameter landscape do not have any global structure similar to the fitness landscape of $f_{23}$.

%% As seen from Figures \ref{fig:c_sphere} and \ref{fig:c_rastrigin}, the shape of adaptive parameter landscapes is also different depending on the rank of each individual.
%% Some previous studies (e.g., \cite{GhoshDCG11,TakahamaS12a,TangDL15}) pointed out that the appropriate parameter pair of $F$ and $\CR$ values depends on the rank of individuals.
%% However, this general guideline has not been verified in the DE community.

As seen from the positions of the star and the circle in Figures \ref{fig:c_sphere} and \ref{fig:c_rastrigin}, a parameter pair actually generated by P-SHADE is far from the best parameter pair.
Ideally, it is desirable that a PAM can generate a parameter pair close to the best parameter pair.
This observation indicates that there is room for improving PAMs in DE.

%This may make 

%Our observation is the first case that support this general heuristic in the DE community.
%% However, this general guideline was not suppurated by any results.
%% Our observation is the first case that support this general heuristic in the DE community.

%% In the DE community, it has been believed that 

%% Kendall $\tau$ values of the 1st and 100th individuals based on their G1 values.
%%    %% }
%%    %% \label{fig:kendall}

%% \begin{figure}[t]
%%    \centering
%%       \includegraphics[width=0.35\textwidth]{graph/kendall/shade_n100_current_to_pbest_1_bin_d20.pdf}
%%    \\
%%    \caption{
%% Kendall $\tau$ values of the 1st and 100th individuals based on their G1 values.
%%    }
%%    \label{fig:kendall}
%% \end{figure}

%Gallagher's Gaussian 101-me Peaks Function

%NFE

%% % indicate that the area with non-zero G1 values is very small like needle-in-haystack landscapes.
%% For example, as seen from results of the 25-th individual in Figure \ref{fig:c_rastrigin}, 

%% rank of individual.

%% Rank

%% Function

%% It is not surprising that the adaptive parameter landscape depends .

%% \begin{observation}
%% Correlation between pal and the landscape of a given problem.
%% \end{observation}

%% \begin{figure}[t]
%%    \centering
%%    \subfloat[$f_1$]{\includegraphics[width=0.23\textwidth]{graph/metaparam/f1_d20.pdf}}
%%    \subfloat[$f_3$]{\includegraphics[width=0.23\textwidth]{graph/metaparam/f3_d20.pdf}}
%%    \caption{
%%      Contour maps of adaptive parameter landscapes in P-SHADE with current-to-$p$best/1 on $f_1$ with $d=20$.
%%    }
%%    \label{fig:shade_h}
%% \end{figure}

\begin{figure*}[t]
  \centering
\subfloat[FDC]{\includegraphics[width=0.32\textwidth]{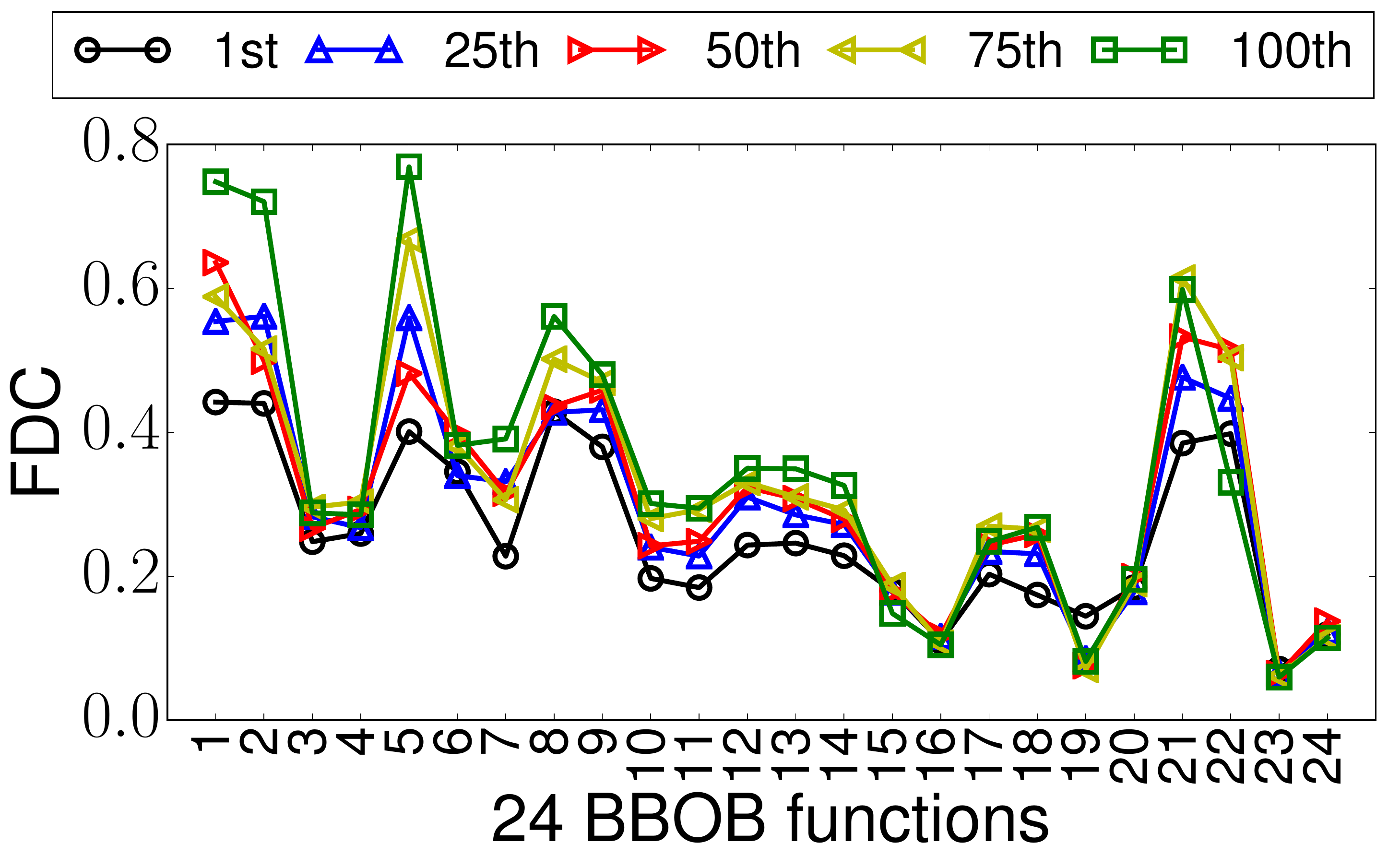}}
\hspace{2mm}
\subfloat[DISP]{\includegraphics[width=0.32\textwidth]{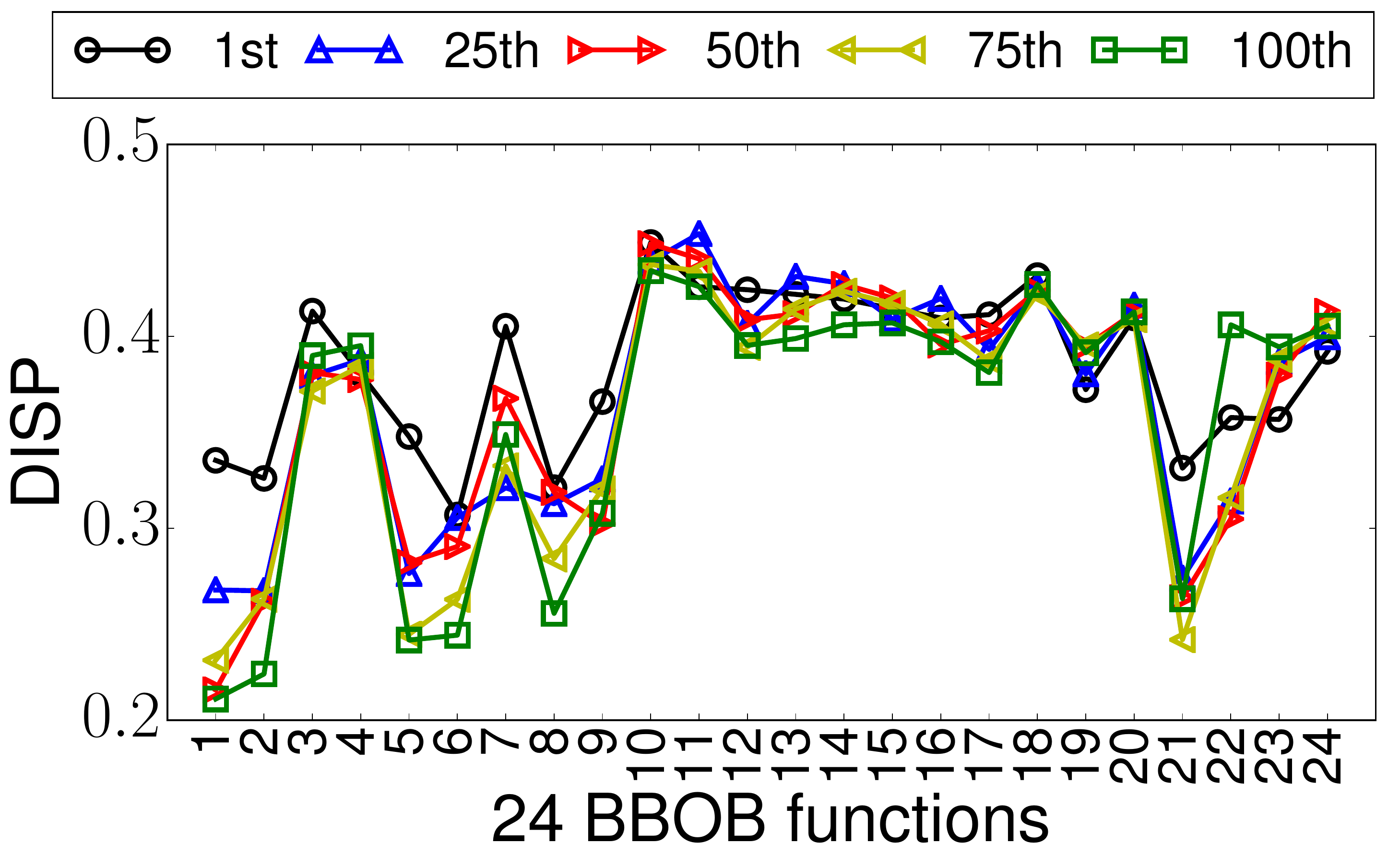}}
\hspace{2mm}
\subfloat[NZR]{\includegraphics[width=0.32\textwidth]{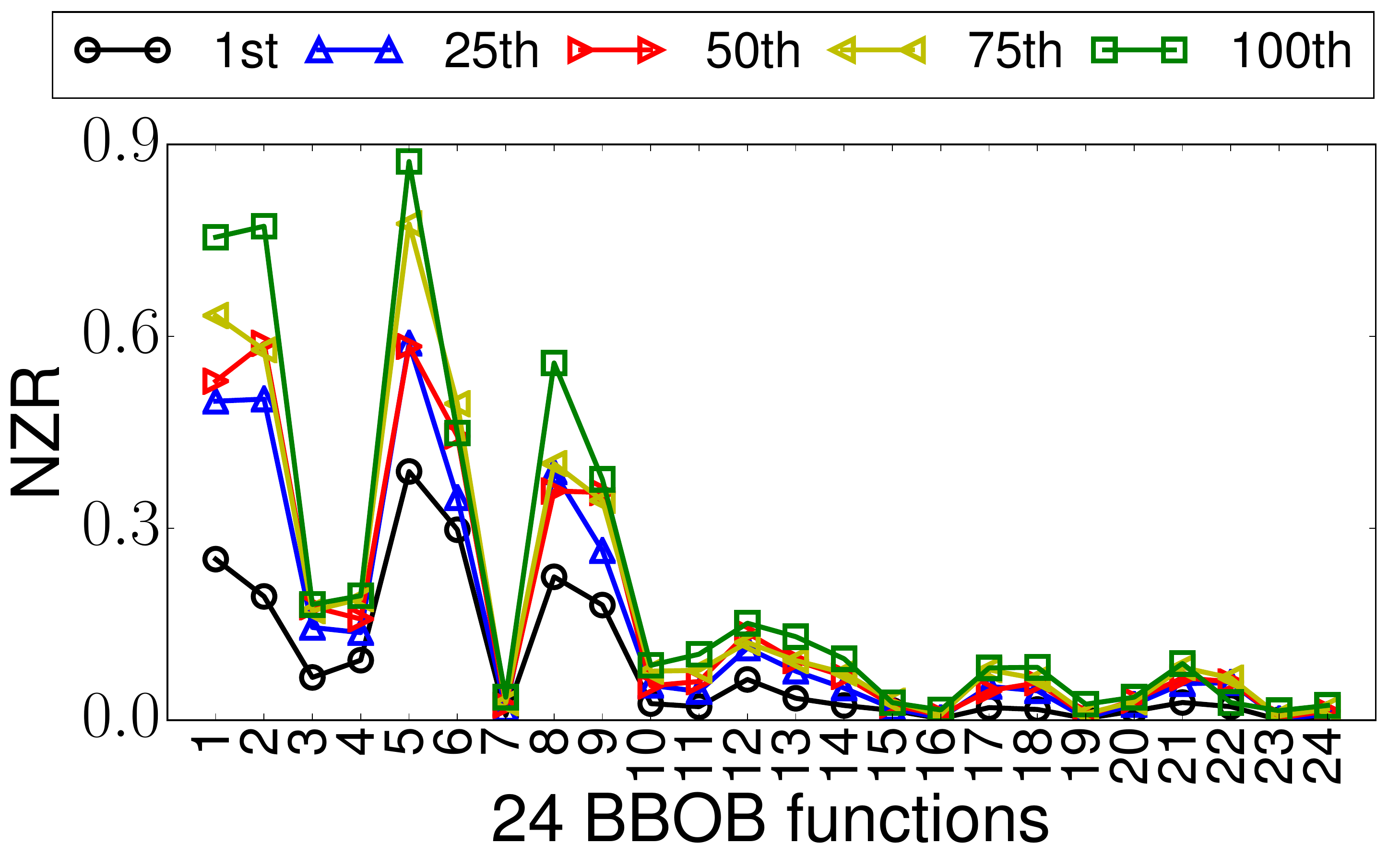}}  
%% \subfloat[rand/1]{\includegraphics[width=0.48\textwidth]{graph/fdc/shade_n100_current_to_pbest_1_bin_d20.pdf}}\\
  %% \subfloat[current-to-$p$best/1]{\includegraphics[width=0.48\textwidth]{graph/fdc/shade_n100_current_to_pbest_1_bin_d20.pdf}}
   \caption{
     Average FDC, DISP, and NZR values of adaptive parameter landscapes in P-SHADE ($d=20$).
   }
   \label{fig:avg_measures}
\end{figure*}

\subsection{Analysis using landscape measures}
\label{sec:fdc}

This section analyzes adaptive parameter landscapes using two representative landscape measures (FDC \cite{JonesF95} and DISP \cite{LunacekW06}) and a non-zero ratio (NZR) measure.
FDC measures the correlation between objective values and the distance to the best solution found (or the optimal solution).
A large FDC value indicates that the corresponding fitness landscape has a strong global structure.
In DISP, first, all solutions are sorted based on their objective values in descending order.
Then, the dispersion of the top $b$ solutions is calculated based on the average pairwise distance between them ($b= \lfloor 0.1 \, m \rfloor$ in this study).
A large DISP value indicates that the corresponding fitness landscape has a multi-funnel.
%Both FDC and DISP quantify the global structure of a given fitness landscape.
Although FDC and DISP were originally proposed for {\em fitness landscape analysis}, they can be extended for {\em parameter landscape analysis} with no significant change, as demonstrated in \cite{HarrisonOE19}.
When using FDC and DISP for {\em adaptive parameter landscape analysis}, ``the objective value'' is replaced with the G1 value, and ``the solution'' is replaced with the parameter pair of $F$ and $\CR$ values.
We did not normalize parameter values since $F \in [0, 1]$ and $\CR \in [0,1]$.
FDC and DISP perform poorly in high-dimensional spaces \cite{MorganG14}, but we address only the two-dimensional space ($F$ and $\CR$).

%When applying FDC to an adaptive parameter landscape of the $i$-th individual at iteration $t$, FDC measures the correlation between G1 values and the distance to the best parameter pair in the $m (=50 \times 50)$ pairs for each individual and each iteration.

%Although FDC is originally proposed for fitness landscape analysis, it can be extended for parameter landscape analysis with no significant change, as demonstrated in \cite{PushakH18,HarrisonOE19}.

%See \cite{JonesF95} and \cite{LunacekW06} for the definisions of FDC and DISP.
%% 
%FDC is one of the most representative fitness landscape measures.

We introduce NZR for analyzing adaptive parameter landscapes.
We do not argue that NZR is one of our contributions since it just counts numbers.
The NZR value of an adaptive parameter landscape in the $i$-th individual at iteration $t$ is given as follows:
\begin{align}
  \label{eqn:nzr}
        {\rm NZR}(\vector{\theta}^t_{i,1}, ..., \vector{\theta}^t_{i,m}) = \frac{1}{m}\left| \left\{\vector{\theta}^t_{i,j} | {\rm G1}(\vector{\theta}^t_{i,j}) >0, j = 1, ..., m \right\}\right|,
\end{align}
where the NZR value is always in the range $[0,1]$.
NZR measures the difficulty in generating a ``successful'' parameter pair of $F$ and $\CR$ values based on the area with non-zero G1 values (see Section \ref{sec:pam} for the definition of ``successful'').
A large NZR value indicates that it is easy to generate a successful parameter pair on the corresponding adaptive parameter landscape.

%% When ${\rm NZR}(\vector{\theta}_1, ..., \vector{\theta}_n) = 1$, it means that all parameter pairs can generate a better trial vector than the corresponding parent vector.
%% When ${\rm NZR}(\vector{\theta}_1, ..., \vector{\theta}_n) = 0$, it means that no parameter pair can generate a better trial vector than the corresponding parent vector.
%The NZR value represents the difficulty in generating a better trial vector than the corresponding parent vector.

   %%   Average FDC, DISP, and NZR values of adaptive parameter landscapes in P-SHADE on the 24 BBOB functions with $d=10$.
   %% }
   %% \label
   
Figure \ref{fig:avg_measures} shows the average FDC, DISP, and NZR values of the 1st, 25th, 50th, 75th, and 100th individuals in P-SHADE at $100$, $1\,000$, $2\,000$, ... function evaluations on the 24 BBOB functions with $d=20$.
Figures S.74--S.82 in the supplementary file show results of the three PAMs on the functions with $d \in \{2, 3, 5, 10, 20, 40\}$.

Figure \ref{fig:avg_measures}(a) shows that all FDC values are non-negative on all functions.
As seen from results on all 24 functions, the worse the individual is, the larger the FDC value.
%This is because an inferior individual has a large area with non-zero G1 values as shown in Figure \ref{fig:avg_measures}(c).
Adaptive parameter landscapes for individuals with similar ranks (e.g., the 50th and 75th individuals) have similar FDC values.
This observation indicates that the global structures of adaptive parameter landscapes can correlate with the rank of individuals.
Some previous studies (e.g., \cite{GhoshDCG11,TakahamaS12a,TangDL15}) gave a rule of thumb that the appropriate parameter pair of $F$ and $\CR$ values may depend on the rank of individuals.
Although this rule of thumb has never been supported by any result, it can be justified by our observation in adaptive parameter landscapes.

%% However, this heuristic rule has never been supported by any result.
%% Our finding can justify 

Results of NZR in Figure \ref{fig:avg_measures}(c) show that generating a successful parameter pair is relatively easy for inferior individuals.
Ali \cite{Ali11} demonstrated that generating a better trial vector than an inferior individual in the population is easy.
We believe that our observation supports a generalization of Ali's observation since it can be applied even to adaptive parameter landscapes. % as shown in Figure \ref{fig:avg_measures}(c).

%This observation based on adaptive parameter landscape analysis is consistent with the previous study \cite{Ali11}, which showed that it is relatively easy for an inferior individual to generate a better trial vector than the best individual.

%Figure \ref{fig:nzr_time} shows that the NZR values of parameter pairs for inferior individual are much higher than those for superior individuals.
%For example, the NZR values of parameter pairs for the 100th individual are high than 0.
%In contrast, the NZR values of parameter pairs for the 1st individual are 0 in many cases.
%% These results indicate that the region that can improve each individual can be tight depending on the quality of an individual.
%% Figure \ref{fig:nzr_avg} show the average values of the NZR values of adaptive parameter landscapes in P-SHADE on the 24 functions with $d=10$.
%% The above-mentioned observation is consistent with Figures \ref{fig:czr_avg}.
%% The NZR value of parameter pairs of the 1st, 25th, 50th, 75th, and 100th individuals are increase as the rank increases.

%a good adaptive parameter landscapes of inferior individuals are 

As shown in Figure \ref{fig:avg_measures}(a), the FDC value is different depending on the function.
While the average FDC values on the three separable and unimodal functions ($f_1$, $f_2$, and $f_5$) are large, those on the multi-modal or nonseparable functions are small, except for $f_{21}$ and $f_{22}$.
Although $f_{21}$ and $f_{22}$ are multi-modal functions, each peak of their fitness landscapes is unimodal.
This property of $f_{21}$ and $f_{22}$ may influence the FDC values of adaptive parameter landscapes.

As shown in Figures \ref{fig:avg_measures}(a) and (b), results of DISP are consistent with the above-mentioned results of FDC in most cases.
Figure \ref{fig:avg_measures}(b) can be viewed as an upside-down version of Figure \ref{fig:avg_measures}(a).
This may be because both FDC and DISP quantify the global structures of fitness landscapes.
An analysis with other landscape measures (e.g., ELA \cite{MersmannBTPWR11} and LON \cite{AdairOM19}) is another future work.

%Altough we showed he results on the 24

Figures S.74--S.82 in the supplementary file show that the results of P-jDE, P-JADE, and P-SHADE for $d \in \{5, 10, 40\}$ are similar to those for $d=20$ (Figure \ref{fig:avg_measures}).
In contrast, results for $d \in \{2, 3\}$ are noisy.
This may be because DE algorithms with PAMs do not work well on such low-dimensional problems as reported in \cite{TanabeF19}.

%% file: conclusion.tex
\section{Conclusion}
\label{sec:conclusion}

%which is a dynamic parameter landscape based on $F$ and $\CR$ values adjusted by a PAM

We have analyzed adaptive parameter landscapes based on $F$ and $\CR$ in PAMs for DE.
We introduced the concept of adaptive parameter landscapes (Section \ref{sec:apl}).
We proposed the method of analyzing adaptive parameter landscapes based on the G1 metric (Sections \ref{sec:g1} and \ref{sec:pal}).
We also examined adaptive parameter landscapes in P-jDE, P-JADE, and P-SHADE on the 24 BBOB functions by using the proposed method (Section \ref{sec:results}).

Our observations in this study can be summarized as follows:

\begin{enumerate}
\renewcommand{\labelenumi}{\roman{enumi})}
\item An adaptive parameter landscape $\mathcal{L}_{a}$ ({\bf {\em NOT the optimal parameter}} $\vector{\theta}^*$) is different depending on the search progress. For example, it is relatively easy to generate successful parameters in an early stage of evolution.
\item $\mathcal{L}_{a}$ ({\bf {\em NOT}} $\vector{\theta}^*$) is significantly influenced by the characteristics of a given problem (e.g., the local/global multimodality).  
\item $\mathcal{L}_{a}$ ({\bf {\em NOT}} $\vector{\theta}^*$) differs depending on the rank of an individual, but $\mathcal{L}_{a}$ of individuals with similar ranks are generally similar.  
\item In most cases, P-jDE, P-JADE, and P-SHADE generate a parameter pair of $F$ and $\CR$ values far from the best parameter pair. This means that there is room for improving PAMs.
%\item Even if the best parameter pairs is used, the best individual is hard to generate a better child.
%\item The characteristics of parameter landscapes and adaptive parameter landscapes are significantly differnt from each other.
%\item Each individual has own parameter landscape.
%
%\item No work provided results that supoort this hypothesis or recet this hypothessis.
\end{enumerate}

We emphasize that our observations about PAMs could not be obtained without analyzing adaptive parameter landscapes.
We believe that our observations can be useful clues to design an efficient PAM.
Overall, we conclude that adaptive parameter landscape analysis can provide important information about PAMs for DE.

%We belive that obtained results are hepful to design an efficient PAM in DE.

%% Correlation between parameter landscapes and adaptive parameter landscapes

%Relation between the performance of DE with a parameter adaptation method and the distance from the best parameter pair.

%Comparing the properties of parameter landscapes and adaptive parameter landscapes is an interesting direction for future work.
Although we examined adaptive parameter landscapes in PAMs for DE, we believe that the proposed analysis method can be applied to PAMs for other evolutionary algorithms, including genetic algorithms and evolution strategies.
Further analysis is needed.

%% Further analysis is needed.
%% Further analysis is needed.
%% Further analysis is needed.